\documentclass[10pt,journal,cspaper,compsoc]{IEEEtran}

\usepackage{dsfont}
\usepackage{algorithmic}
\usepackage{algorithm}
\usepackage{amsmath}
\usepackage{amssymb}
\usepackage{makeidx}
\usepackage{graphicx}
\usepackage{booktabs}
\usepackage{multirow}
\usepackage{subfigure}
\usepackage{tabularx}

\newcommand*{\bv}[1]{\ensuremath \boldsymbol{#1}}
\newcommand*{\mcal}[1]{\ensuremath \mathcal{#1}}

\newcommand*{\mth}{\mcal{M}_{\theta}}
\newcommand*{\Xb}{\bv{X}}
\newcommand*{\Vb}{\bv{V}}
\newcommand*{\fb}{\bv{f}}
\newcommand*{\xb}{\bv{x}}
\newcommand*{\yb}{\bv{y}}
\newcommand*{\thb}{\bv{\theta}}
\ifCLASSOPTIONcompsoc
\else
\fi
\ifCLASSINFOpdf
\else
\fi
\begin{document}
\title{Fast Approximate Bayesian Computation
for Estimating Parameters in Differential Equations
}

\author{Sanmitra~Ghosh, %~\IEEEmembership{Member,~IEEE,}
        Srinandan~Dasmahapatra, %~\IEEEmembership{Member,~IEEE,}
        and~Koushik~Maharatna,~\IEEEmembership{Member,~IEEE}% <-this % stops a space
\IEEEcompsocitemizethanks{\IEEEcompsocthanksitem S. Ghosh, S. Dasmahapatra and K. Maharatna are with the Department
of Electronics and Computer Science, University of Southampton, Southampton,
SO17 1BJ.\protect\\
E-mail: \{sg5g10, sd, km3\}@ecs.soton.ac.uk
}% <-this % stops a space
\thanks{}}

\IEEEcompsoctitleabstractindextext{%
\begin{abstract}
%\boldmath
Approximate Bayesian computation (ABC) using a sequential Monte Carlo method provides a comprehensive platform for parameter estimation, model selection and sensitivity analysis in differential equations. However, this method, like other Monte Carlo methods, incurs a significant computational cost as it requires explicit numerical integration of differential equations to carry out inference. In this paper we propose a novel method for circumventing the requirement of explicit integration by using derivatives of Gaussian processes to smooth the observations from which parameters are estimated.  We evaluate our methods using synthetic data generated from model biological systems described by ordinary and delay differential equations. Upon comparing the performance of our method to existing ABC techniques, we demonstrate that it produces comparably reliable parameter estimates at a significantly reduced execution time. 
   
\end{abstract}
% not want either math or citations to appear in the abstract.

% Note that keywords are not normally used for peer review papers.
}

\maketitle

% is not selected <OR> if conference mode is selected - because compsoc
% conference papers position the abstract like regular (non-compsoc)
\IEEEdisplaynotcompsoctitleabstractindextext

% \fi
\IEEEpeerreviewmaketitle

%\section{Introduction}
% Computer Society journal papers do something a tad strange with the very
% first section heading (almost always called "Introduction"). They place it
% ABOVE the main text! IEEEtran.cls currently does not do this for you.
% However, You can achieve this effect by making LaTeX jump through some
% hoops via something like:
%
\ifCLASSOPTIONcompsoc
  \noindent\raisebox{2\baselineskip}[0pt][0pt]%
  {\parbox{\columnwidth}{\section{Introduction}\label{sec:introduction}%
  \global\everypar=\everypar}}%
  \vspace{-1\baselineskip}\vspace{-\parskip}\par
\else
  \section{Introduction}\label{sec:introduction}\par
\fi
%
% Admittedly, this is a hack and may well be fragile, but seems to do the
% trick for me. Note the need to keep any \label that may be used right
% after \section in the above as the hack puts \section within a raised box.

\IEEEPARstart{T}{he} time evolution of the variables modelled in a variety of science and engineering branches are often described by ordinary differential equations that are characterised by model structure -- the functions of the dynamical variables -- and model parameters. The task of estimating these parameters from experimental observations is thus of paramount importance. It is also necessary in some cases to choose the most appropriate among competing models that describe the observations. For parameter estimation and model selection, statistical and pattern recognition techniques built upon a Bayesian framework have been shown to work extremely well for complex non-linear ordinary differential equations in \cite{Calderhead},\cite{dondelinger2013ode},\cite{Barber2014} and \cite{Vyshemirsky2008}.
 
To apply Bayesian techniques we need to integrate marginal likelihoods, which can be computationally intractable in non-linear differential equation models. For this reason some form of approximation such as Monte Carlo integration is generally preferred for parameter inference.  Approximate Bayesian computation (ABC) based on sequential Monte Carlo (SMC) is one such approximate inference technique that has been applied to different classes of dynamical systems described by deterministic or stochastic differential equations for both parameter estimation and model selection in \cite{Toni2009}. The ABC-SMC algorithm has been shown to work well for the examples considered in \cite{Toni2009}.  ABC-SMC produces reliable estimates of parameters and has been used to discriminate between a set of candidate models using Bayesian model selection criteria. Moreover,  ABC-SMC enables the calculation of parameter sensitivities \cite{Toni2009}. 

ABC methods prove to be most useful for large models with complex likelihood surfaces that are difficult to evaluate.  The operating principle of ABC methods lies in replacing the evaluation of likelihoods with a simulation based procedure for inference, by using a generative model $\mth$ with parameters $\theta$ drawn from a prior distribution $\pi(\theta)$ to simulate observations $Y^s\sim\mth$ that are compared with the observed data $Y^d$.   If the likelihood $p(Y^d|\mcal{M}_\theta)$ of observed data $Y^d$ is intractable or infeasible to compute, then we can use the ABC algorithm to obtain samples from the following modified posterior density
 \begin{equation}\label{eq:ABCBayes}
\begin{aligned}
& p_\epsilon(\theta,Y^s|Y^d) =\\
&\frac{\mathds{1}(\Delta(Y^d,Y^s)\leq \epsilon)(Y^s\sim\mth)\pi(\theta)}
{\int_{\theta}{\int_{Y^s}{\mathds{1}(\Delta(Y^d,Y^s)\leq \epsilon)(Y^s\sim\mth)\pi(\theta)d\theta dY^s}}}
\end{aligned}
\end{equation}
where $\epsilon>0$ is a tolerance level, $\Delta$ is a distance function, $\mathds{1}$ is the indicator function and $p_\epsilon(\theta,Y^s|Y^d) =p(\theta,Y^s|\Delta(Y^d,Y^s)\leq\epsilon)$. 
A good (enough) approximation of the true marginal posterior distribution is obtained when the distance $\Delta(Y^d,Y^s)$ is within a predetermined small tolerance $\epsilon$, \emph{i.e.,}
\begin{equation}
p_\epsilon(\theta|Y^d)=\int_{Y^s}{p_\epsilon(\theta,Y^s|Y^d)dY^s}\approx p(\theta|Y^d)
\end{equation}
Since ABC (including ABC-SMC) requires the generation of a number of simulated observations $Y^s \sim\mth$, the generation of observations could be a computationally expensive process.  Thus although ABC-SMC mitigates the intractability of evaluating the likelihood function through simulation, repeated simulation from complex models for inference can itself be burdensome. For the case of dynamical systems such simulations require explicit numerical solutions of non-linear differential equations. Thus, despite its attractive features the ABC-SMC algorithm suffers from a major drawback rooted in its computational burden for inference in differential equations. In particular, the acceptance criterion $\mathds{1}(\Delta(Y^d,Y^s)\leq \epsilon)$ can lead to the generation of many unused simulations $Y^s\sim\mth$, and  various methods  have been proposed in \cite{Filippi2013} to improve the acceptance rate and reduce the run-time of the algorithm.

In this paper we propose an alternate method of speeding up the ABC-SMC algorithm for parameter estimation in deterministic models described by ordinary differential equations (ODE) or delay differential equations (DDE) by reducing the time incurred in simulation. We achieve this speedup by: (i) completely circumventing the process of integrating the differential equation by operating on the derivative space and (ii) by smoothing the derivatives using Gaussian processes (GP). It should be noted that using Gaussian processes as functional emulators in the derivative space, as a concept, has been proposed in \cite{Calderhead},\cite{dondelinger2013ode} for speeding up parameter estimation in deterministic differential equations.  For parameter estimation, GP-based gradient matching has been used for ODEs and DDEs using a population Monte Carlo sampling \cite{Calderhead}; an adaptive variant of this approach is proposed for ODEs  \cite{dondelinger2013ode}. See \cite{Barber2014} for a review and comparison between these approaches. The novelty of our proposed method is the fusion of GP regression with ABC-SMC.  Our algorithm for fast parameter estimation can be easily incorporated into methods for model selection and recovering parameter sensitivities for deterministic differential equations.  

%Thus apart from providing a fast Bayesian estimate of parameters as in \cite{Calderhead,dondelinger2013ode} we can further embed certain qualitative features of the dynamical system, whose parameters we intend to estimate, within the distance functions. For example, when considering a dynamical system with a periodic orbit, we can embed its periodicity by creating a distance function based on its spectral signature. Thus with the fusion of GPR with ABC-SMC, we are able to propose a single holistic algorithm for fast parameter estimation that can be easily incorporated into the method for model selection and recovering parameter sensitivities for deterministic differential equations.

This paper is organised as follows: in section \ref{sec:ABCbasics} we will introduce the ABC-SMC algorithm for parameter estimation in differential equations. In section \ref{sec:gradient} we will show how the algorithm can be sped up by circumventing the need for the numerical solution of the differential equations. In section \ref{sec:GP} we will introduce Gaussian processes for function estimation and subsequently in section \ref{sec:ABC-SMC} we will use GPs within the ABC-SMC to operate on the derivative space. In section \ref{sec:eval} we will compare the performance of our proposed modification with the ABC-SMC  algorithm \cite{Toni2009} and also with its improved variant proposed in \cite{Filippi2013}. In conclusion, we summarise our achievements in this paper and give some indication of future work in section \ref{sec:conc}.


\section{The basic ABC framework}\label{sec:ABCbasics}
%\IEEEpubidadjcol
 ABC methods generally have the following algorithmic form:
\begin{enumerate}
\item Sample a candidate parameter vector $\theta$ from a prior distribution $\pi(\theta)$ and for each $\theta\sim\pi(\theta)$, simulate a dataset $Y^s\sim\mth$ from a generative model $\mth$.
\item Compute a distance $\Delta(Y^s,Y^d)$ between the simulated dataset, $Y^s$ and the experimental data $Y^d$. If $\Delta(Y^d,Y^s)\leq \epsilon$, where $\epsilon\geq 0$ is the error tolerance of accepted solutions, then accept $\theta$ and reject otherwise. 
\end{enumerate}
These scheme is repeated until $N$ parameter values are accepted, which represent a sample from the approximate posterior distribution $p_\epsilon(\theta|Y^d)$. Exact posterior can be obtained from this scheme when $\epsilon= 0$.

\subsection{ABC-SMC for parameter estimation in ODE}

If the prior distribution is very different from the posterior distribution, the basic ABC framework is very inefficient as it spends a considerable amount of time sampling from areas of low likelihood in parameter space, which makes the acceptance rate extremely low. In order to improve upon poor acceptance rates and facilitate exploration of the parameter space, ABC algorithms based on the SMC sampling method were proposed in \cite{sisson2007sequential}, \cite{DelMoral2006} and \cite{SC2009} and sequential importance sampling (SIS) in \cite{Toni2009}, \cite{Beaumont2009}. \cite{Toni2009} applied the ABC algorithm based on SIS for parameter estimation and model selection for a variety of dynamical systems including non-linear ODEs and DDEs, which will also be the focus in this paper.  Although all the variants of ABC algorithms that come under the SMC category can potentially be used for inference in dynamical systems, we will specifically focus our attention on the ABC approach as adopted in \cite{Toni2009}. 

We shall apply ABC-SMC to models of the evolution of state $\bv{X}(t)=(X_1(t),\ldots,X_K(t)$ that are governed by ODEs or DDEs $\frac{d\Xb(t)}{dt}=f({\bv{X}(t - t_d)},\thb)$, where $t_d$ stands for the time delay in DDEs, with $t_d=0$ for ODEs, and $\thb$ is a vector of parameter values.  We express the integrated solution of the differential equations $\Xb(t, \Xb_{\rm{in}};\thb)$ as a map $\psi_t(\Xb_{\rm{in}};\thb)$ that generates state trajectories $\Xb(t)$ given a set of parameters $\thb$ and initial conditions $\Xb_{\rm{in}}\triangleq\Xb(t_d\leq t\leq 0)$.  To generate the samples $Y^s$, we add Gaussian noise $\bv{\eta}\sim\mcal{N}(\bv{0},\sigma^2\mathbb{I}_{K})$, where $\mathbb{I}_{K}$ is a $K\times K$ identity matrix to the solutions $\Xb(t, \Xb_{\rm{in}};\thb)$ to create the generative model $\mth$:
\begin{equation}\label{eq:gen}
(Y^s\sim\mth) \Leftrightarrow Y^s = \Xb(t, \Xb_{\rm{in}};\thb).
\end{equation}
to be used in the ABC framework.

A collection of parameter values, called particles $\thb$ are sampled from the prior $\pi(\thb)$ to instantiate the generative model $\mth$. To decide whether a particular choice of parameters $\thb\sim\pi(\thb)$ is accepted, we need to compare if the simulated trajectory $Y^s\sim\mth$ is within a tolerance level $\epsilon$ of the observed trajectory $Y^d$, for which we introduce the distance function   
\begin{equation}\label{eq:distance}
\begin{aligned}
\Delta(Y^d,Y^s)=&\sum_{i=1}^L\sum_{k=1}^K (Y^d_k(t_i)-X_k(t_i)-\eta_k)^2,
\end{aligned}
\end{equation}
%where the comparison is performed at $L$ time instances indexed by $t$.
where we assumed that the data were collected, and the state evaluated, at integer time points $t^L\triangleq \{t_i\}_{i=1,\ldots,L}$.  Note that for dynamical systems such distance functions are generally built by considering the entire time-series data instead of some sufficient statistics.

The sequential stage of this algorithm involves replacing a single tolerance value $\epsilon$ by a sequence of tolerance values $\epsilon_{\tau}$, where $\tau=0,\ldots,S_{\scriptstyle MC}$ denotes the sequential steps, and $\epsilon_{\tau}>\epsilon_{\tau+1}$.  The particles $\thb_{\tau}$ are indexed by $\tau$ labelling the tolerance   level, and are sampled from the posterior distribution obtained from the previous sequential step, thus introducing a step-wise procedure for generating parameters from a sequence of more informative distributions, starting at $\tau=0$ with the prior distribution $\pi(\thb)$.  To accept or reject the sampled particles for sequence index $\tau$, the generated trajectories from the model with parameters $\thb_{\tau}$ must be closer to the observed data $Y^d$ than those generated from the model with parameters $\thb_{\tau-1}$ in step $\tau-1$.  The generative mechanism for the particles $\thb_{\tau}$ differs from that of $\thb_{\tau-1}$ in that they are sampled from the $N$ particles $\{\thb_{\tau-1}^{(i)}\}_{i=1,\ldots,N}$ with importance weights \cite{Toni2009} $w_{\tau-1}$ and each $\thb^*\sim\{\thb_{\tau-1}^{(i)}\}_{i=1,\ldots,N}$ is perturbed by a perturbation kernel $K_{\tau}(\thb | \thb^*)$ \cite{Toni2009}. 

For each $\tau=0,\ldots,S_{\scriptstyle MC}$, the $N$ particles meeting the acceptance criterion $\Delta(Y^s,Y^d)\leq \epsilon_{\tau}$ represent a point-wise approximation for the posterior distribution over the parameter values:
\begin{equation}\label{eq:ImpWght}
p_{\epsilon_{\tau}}(\thb|Y^d)\approx\frac{1}{N}\sum_{i=1}^Nw^{(i)}_{\tau}\delta (\thb-\thb^{(i)}_{\tau}), 
\end{equation}
where $w^{(i)}_{\tau}$ is the importance weight of the particle $i$ \cite{Toni2009} and $\delta (\thb-\thb^{(i)})$ is a product of Dirac delta functions, one for each component of $\thb$.  For a small value of $\epsilon_{S_{\scriptstyle MC}}$ the final collection of particles should be a good point-wise approximation to the true posterior distribution. The ABC-SMC algorithm is listed in Algorithm 1. 

\algsetup{
linenosize=\small,
linenodelimiter=.
}
\begin{algorithm}
\caption{ABC-SMC as proposed in \cite{Toni2009}}
\label{alg1}
\begin{algorithmic}[1]
 \STATE Given $Y^d$, $\pi(\thb)$, $\mth$.
\STATE Initialise $\epsilon_{\tau}>0$, $\tau=0,\ldots,S_{\scriptstyle MC}$, $\epsilon_{\tau}>\epsilon_{\tau+1}$. Set $\tau=0$.
\STATE Set $i=1$.
\IF {$\tau=0$}
\STATE sample $\thb^{**}$ independently from $\pi(\thb)$: \\ $\thb^{**}\sim\pi(\thb)$
\ELSE  
\STATE  from the previous population $\{{\thb^{(i)}_{\tau-1}}\}_{i=1,\ldots,N}$\\sample $\thb^{*}\sim\{{\thb^{(i)}_{\tau-1}}\}_{i=1,\ldots,N}$ with associated normalized weights $w^{*}_{\tau-1}$ and use the perturbation kernel $K_\tau(\thb|\thb^*)$ to produce $\thb^{**}\sim K_\tau(\thb|\thb^*)$.
\ENDIF
\IF {$\pi(\thb^{**})=0$}
\STATE go to $4$. 
\ELSE  
\STATE Simulate a candidate dataset $Y^s$ from the model $\mth$ with parameter $\thb^{**}$: $Y^s\sim\mth\vert_{\theta\leftarrow\thb^{**}}$.
\ENDIF
\IF {$\Delta(Y^d,Y^s) \geq\epsilon_{\tau}$}
\STATE go to $4$. 
\ELSE  
\STATE Set $\thb^{(i)}_{\tau}\leftarrow\thb^{**}$ and calculate the weight for particle $\thb^{(i)}_{\tau}$,
\[ w^{(i)}_{\tau} = \left\{ 
  \begin{array}{cc}
    $1$, & \quad \mbox{if $\tau=0$} \\
    \frac{\displaystyle\strut\pi(\thb^{(i)}_{\tau})}{\displaystyle\sum_{j=1}^{N} w^{(j)}_{\tau-1}K_\tau(\thb^{(i)}_{\tau}\vert\thb^{(j)}_{\tau-1})}, & \quad \mbox{if $\tau>0$}
  \end{array} \right.\]
\ENDIF
\IF {$i<N$}
\STATE Set $i\leftarrow i+1$ and go to $4$. 
\ELSE  
\STATE Normalise the weights.
\ENDIF
\IF {$\tau<S_{\scriptstyle MC}$}
\STATE Set $\tau\leftarrow \tau+1$ and go to $3$. 
\ELSE  
\RETURN particles $\theta^{(i)}_{S_{\scriptstyle MC}}$ at $\tau=S_{\scriptstyle MC}$.
\ENDIF
\end{algorithmic}
\end{algorithm}

One way of speeding up the ABC-SMC algorithm is by increasing the acceptance rate. To this end a range of perturbation kernels were proposed in \cite{Filippi2013} that result in a noticeable change in terms of acceptance rates and run-time. However, the major computational burden stems from the numerical integration of the differential equation and thus a faster simulation method within the ABC-SMC is likely to speed up this algorithm possibly more than what is gained by clever choice of perturbation kernels.

\section{Gradient based parameter estimation in differential equations}\label{sec:gradient}

We have mentioned previously that the computational bottleneck stems from the explicit integration carried out in each simulation step. In order to avoid the integration one could essentially use a gradient based estimation. If the temporal variations in observations $Y^d(t)$ are believed to be less smooth than the underlying state evolution that is modelled by differential equations, we shall introduce the target state variable $\widehat{\Xb}(t)$ to be the smoothed version $\widehat{\Xb}(t)\triangleq\mcal{S}(Y^d)$ of the observations.  Here, $\mcal{S}$ represents any smoothing procedure, and we will use Gaussian Process (GP) regression to perform the smoothing below. In the ABC framework, we shall accept the trajectories $\Xb(t)$ from the model $\mth$ (see (\ref{eq:gen})) if they are close to $\widehat{\Xb}(t)$.  Once we have $\widehat{\Xb}(t)$ we can compute its numerical derivative to obtain the empirical vector field $\Vb^d(t)$ of the dynamical system $\mth$:
\begin{equation}\label{eq:velocity}
\mbox{for } \widehat{\Xb}(t)\triangleq\mcal{S}(Y^d), \Vb^d(t)\triangleq\frac{d}{dt}\widehat{\Xb}(t).
\end{equation}
In addition, while the left hand side of the ODE $\frac{d}{dt}\Xb(t)=\fb({\Xb(t)},\thb)$ is estimated by the empirical derivative $\Vb^d(t)$, it should be matched by the model vector field $\fb(\Xb(t), \thb)$ on the right hand side, when evaluated on the smoothed state data $\fb(\Xb(t)=\widehat{\Xb}(t), \thb)$.  

Upon introducing a new distance measure between $\Vb^d(t)$ obtained from the smoothing and $\fb(\widehat{\Xb}(t),\thb)$ obtained from the vector field we can eliminate the original distance metric for ABC-SMC between observed, $Y^d$, and the simulated, $Y^s\sim\mth$, trajectories, thus unburdening ABC-SMC of ODE integration at each simulation step. The gradient based method was first suggested in \cite{varah1982spline} where a spline-based smoothing was used to denoise the observed data. In this method a cost function was built using the distance metric in derivative space and optimisation was used to minimise this cost function in order to obtain point estimates. Recent developments of this methods are described in \cite{Ramsay2007}. All these approaches suffer from similar problems of using additional regularisation parameters for smoothing  and often the estimates are sub-optimal point estimates.  Although porting the derivative based distance within an ABC scheme alleviates the computational bottleneck, this approach suffers from an inherent shortcoming that is rooted in obtaining a numerical derivative as this might lead to information loss.

In our approach, we replace the numerical differentiation with a zero mean Gaussian Process (GP) prior on the state $\Xb(t)$ given by
\begin{equation}\label{eq:GP prior}
 p(\Xb(t)|\phi)\sim \mathcal{GP}(0,K(t,t';\phi)),
\end{equation}
 where $K(t,t^{'};\phi)$ denotes a covariance function with hyperparameters $\phi$. Once such a prior is established then Gaussian Process regression techniques can be applied to estimate both the state vector $\widehat{\Xb}(t)$ and also the derivative process $\Vb^d(t)$. Using GP regression the derivative process can be inferred within a probabilistic framework. Hence, we propose to use a distance function in the derivative space where the state $\widehat{\Xb}(t)$ and derivative $\Vb^d(t)$ is modelled using GP regression,within the ABC-SMC algorithm. In this way our proposed method is based on the GP construction in the derivative space as in \cite{Calderhead} and \cite{Barber2014}, combined with the ABC-SMC algorithm as proposed in \cite{Toni2009}. Next, we will briefly introduce GP regression and will apply this to the ABC-SMC algorithm.
 
\section{Gaussian processes} \label{sec:GP}

For real-valued functions $f:\mathcal{A}\rightarrow \mathbb{R}$ of one or more input variables defined over $\mcal{A}$, a Gaussian process (GP) is a Bayesian non-parametric model that specifies a distribution $P(f)$    \cite{O'Hagan1978,Mackay1998,Neal1998,Rasmussen2006} characterised by a mean function $\mu(x)$ and a covariance function or kernel, $K(x,x';\phi)$:
\begin{equation}
f(x) \sim \mathcal{GP}(\mu(x),K(x,x';\phi)),
\end{equation}
where $\phi$ are hyperparameters. For example, the squared exponential covariance function, which we use below, is given by
\begin{equation}\label{eq:SEiso}
K_{SE}(x,x')=\sigma^2_{kern}\exp\left(-\frac{1}{2}\frac{(x-x')^2}{l^2}\right),
\end{equation}
with hyperparameters $\sigma^2_{kern}$ and $l^2$ (variance and characteristic length-scale). 

For a finite number ($n$) of inputs $\xb^*=(x_1, \ldots. x_n)$, $x_i \in \mathcal{A}$ and for $f(x)\sim\mcal{GP}(\mu(x),K(x,x';\phi))$, the $n$-dimensional vector of function values evaluated at $n$ points $\fb(\xb^*)\triangleq\left(f({x_1}),...,f({x_n})\right)$  is a random vector drawn from a multivariate Gaussian distribution:  
\begin{equation}
p(\fb(\xb^*)|\xb^*) = \mathcal{N}(\mu(\xb^{*}),K(\xb^{*},\xb'^{*})).
\end{equation}

For performing regression, observations $\yb(\xb)\triangleq (y(x_1), \ldots, y(x_L)$ at $L$ training points $\xb=(x_1,\ldots, x_L)$ are fit to function $f$ evaluated at $\xb$:
\begin{equation}\label{eq:Noisyobs}
\yb(\xb)=\fb(\xb)+\bv{\eta}, 
\end{equation} 
as in (\ref{eq:gen}).  Given the training data $(x_i, y(x_i))$, $i=1,\ldots,L$ the conditional predictive distribution of the function $\fb^*\triangleq\fb(\xb^*)$ evaluated at the test points $\xb^*$ is a Gaussian with mean and variance given by \cite{Rasmussen2006}
\begin{equation}
\begin{aligned}
E[\fb^*|\yb,\xb,\xb^*,\Phi] & =\mu(\fb^{*})+ \\
& [K(\xb^{*},\xb)(K(\xb,\xb)+\sigma^2 \mathbb{I})^{-1}\\
& \times(\yb-\mu(\fb^*))],\\ 
Var[\fb^*|\yb, \xb, \xb^*,\Phi] & = K(\xb^{*},\xb^{*})-\\
& [ K(\xb^{*},\xb)(K(\xb,\xb)+\sigma^2 \mathbb{I})^{-1}\\
& \times K(\xb, \xb^{*})],
\end{aligned}
\end{equation}
 where $\Phi=\left\{\phi,\sigma\right\}$.

The hyperparameters $\Phi$ are inferred as point estimates by optimising the logarithm of the marginal likelihood, with a zero mean assumption \cite{Rasmussen2006}:
\begin{equation}
\begin{aligned}
\log p(\yb|\Phi) & =\log \int{p(\yb|\fb,\Phi)p(\fb|\Phi)d\fb}\\
&=-\frac{1}{2}\yb^{T}(K(\xb,\xb)+\sigma^2 \mathbb{I})^{-1}\yb\\ 
&-\frac{1}{2}\log \left|K(\xb,\xb)+\sigma^2 \mathbb{I}\right|
-\frac{L}{2}\log (2\pi).
\end{aligned}
\end{equation}

\subsection{Derivative Gaussian processes} \label{sec:DGP}

Since differentiation is a linear operator, the derivative of a Gaussian process is another Gaussian process \cite{Solak2002}. This makes it possible to include derivative observations in the GP model, or to compute predictions about derivatives. We have
\begin{equation}\label {eq:DGP_mean}
E\left[\frac{\partial f(x)}{\partial x}\right]=\frac{\partial E\left[ f(x)\right] }{\partial x}.
\end{equation}
And likewise the covariance between partial derivative and a function value can be written as
\begin{equation}\label {eq:derivative_cov}
K \left(\frac{\partial f(x)}{\partial x}, f(x^*)\right)=\frac{\partial }{\partial x} K \left(x,x^{*}\right),
\end{equation}
and the covariance between partial derivatives follows
\begin{equation}
K \left(\frac{\partial f(x)}{\partial x},\frac{\partial f(x^*)}{\partial x^*}\right)=\frac{\partial^2 }{\partial x \partial x^*} K \left(x,x^*\right).\end{equation}
For example considering the squared exponential covariance function given in (\ref{eq:SEiso}),
we can write the covariance between partial derivative and a function value as
\begin{equation}\label{eq:dervSEiso}
K \left(\frac{\partial f(x)}{\partial x}, f(x^*)\right)=-\frac{(x-x^{*})}{l^2}K(x,x^*)
\end{equation}
The conditional distribution of the derivative of a test function $\frac{\partial f(x^*)}{\partial x^*}$ having observed the output $(x,y)=(x,f(x))$ of a noisy function $f$ is given as
\begin{equation}
P(\frac{\partial f(x^*)}{\partial x^*}|f(x))=\mathcal{N}(\mathbf{m},\boldsymbol{\Sigma}),
\end{equation}
where we have the mean function $\mathbf{m}$ and covariance function $\mathbf{\Sigma}$ (obtained by using (\ref {eq:DGP_mean}) and (\ref {eq:derivative_cov}) and considering the prior mean of the test and training function to be zero)  given as
\begin{equation}\label{eq:DGP_inf}
\begin{aligned}
&\mathbf{m} = \frac{\partial K \left(x^*,x\right)}{\partial x^*} \left[ K(x,x)+\sigma^2 \mathbb{I}\right]^{-1}f(x) ,\\
& \boldsymbol{\Sigma}= \frac{\partial^2 K \left(x^*,x^*\right)}{\partial x^*\partial x^*} \\&- \frac{\partial K \left(x^*,x\right)}{\partial x^*}\left[ K(x,x)+\sigma^2 \mathbb{I}\right]^{-1}\frac{\partial K \left(x,x^*\right)}{\partial x^*}.
\end{aligned}
\end{equation}

\section{ABC-SMC with Derivative GP} \label{sec:ABC-SMC}

In this section we apply the machinery reviewed in the previous sections to the task of inferring parameters in differential equation models whose solution is the state trajectory $\Xb(t)$.  If we assign a GP prior to the state evaluated at time points $t^L\triangleq \{t_i\}_{i=1,\ldots,L}$, then the set of values of the state $\Xb(t^L)$ takes on a Gaussian \bf{prior} \normalfont distribution: 
\begin{equation}
p(\Xb(t^L)|t^L)=\mathcal{N}(\Xb(t^L)|0,K(t^L,t^L)).
\end{equation}
The modelling task is to represent the experimental data as (\ref{eq:gen}) $Y^d=\{\Xb(t^L) + \bv{\eta}^L\}$ where $\bv{\eta}^L$ refers to $L$ i.i.d. samples from $\mcal{N}(0,\sigma^2\mathbb{I})$ and $\Xb(t)$ satisfies a differential equation. We can use GP regression to obtain the expectation and variance of the posterior (given training data $Y^d$ at $t^L$) state $\Xb(t^*)$ for some test input time point $t^*$ as in Section \ref{sec:GP} \cite{Rasmussen2006}:
\begin{equation}\label{eq:GP state}
\begin{aligned}
&E[\Xb(t^{*})|Y^d] = K(t^*,t^L)(K(t^L,t^L)+\sigma^2 \mathbb{I})^{-1}Y^d,\\
 &Var[\Xb^{*}]= K(t^*,t^*) -K(t^*,t)(K(t,t)+\sigma^2 \mathbb{I})^{-1}K(t,t^*).\end{aligned} 
\end{equation}
This expected posterior state variable $\Xb(t^*)$ for arbitrary choice of $t^*$ models the smoothed evolution of the state $\widehat{\Xb}(t)$ introduced above, and where it is assumed that observational noise accounts for deviations from the smoothed time course.  The smoothed state estimation enables us to compute the velocity field using the derivative GP (as in Section \ref{sec:DGP}): 
\begin{equation}\label{eq:GP state derivative}
E[\frac{d}{dt}{\Xb}] = \frac{\partial K(t,t)}{\partial t}( K(t,t) +\sigma^2 \mathbb{I})^{-1}E[\Xb].
\end{equation}
This completes the procedure for deriving the empirical velocity field (\ref{eq:velocity}) $\Vb^d(t) = E[\frac{d}{dt}{\Xb}]$.

To apply this derivative process within the ABC framework we need to define a distance metric $ \Delta(\Vb^d(t),\fb(\widehat{\Xb}(t),\thb))$ between the smoothed velocity field derived from the observed data, and the velocity field postulated in a differential equation model, where the expected state estimation $\widehat{\Xb}(t)$ has been substituted for the state variable. Hence our proposed fast alternative ABC-SMC based on GP gradient distance (GP-ABC-SMC) works as follows:
\begin{enumerate}
\item Having given data $Y^d$ as a noisy observation of the true state variable $\Xb(t)$, assign a GP prior on $\Xb(t)$ using (\ref{eq:GP prior}) and choose a covariance function, with some unknown hyperparameters, needed to define the GP prior.
\item Learn the hyperparameters of the covariance function from  the original noisy experimental data $Y^d$ using maximum likelihood estimation and then run GP regression to obtain an estimation of the smoothed state evolution $\widehat{\Xb}(t)=E[\Xb]$ using (\ref {eq:GP state}) and the experimental time points $t$ as both the training and test input points.
\item Construct the first derivative of the covariance matrix and estimate the derivative process  $\Vb^d(t) = E[\frac{d}{dt}{\Xb}|_{\Xb=\widehat{\Xb}}]$ using (\ref {eq:GP state derivative}).
\item Run the ABC-SMC algorithm with a modified distance metric $ \Delta(\Vb^d(t),\fb(\widehat{\Xb}(t),\thb))\leq \epsilon_{\tau}$ for tolerance schedule $\{\epsilon_{\tau}\}$, where at each simulation step the simulated data $Y^s=\fb(\widehat{\Xb}(t),\thb)$ is generated by evaluating the velocity field on the right hand side of the differential equation.  This yields the posterior distribution of the parameters $p(\thb|Y^d)$ (\ref{eq:ImpWght}).

%\footnote {Note that the generative model $f(t,\theta)$ and the right hand side of the differential equation $f(X,\theta)$ denotes the same thing in ABC context.}  
\end{enumerate}

Note that no explicit solution of differential equation is required to generate the simulated data within the iterations of the GP-ABC-SMC algorithm. Also note the fact that, in order to run the GP-ABC-SMC algorithm, no knowledge of the initial condition is required.

For the sake of conformity with the ABC terminologies, we will persist in using the terms $Y^d$ for observed data and $Y^s$ for simulated data, as before.  However, within the context of GP-ABC-SMC algorithm, observed and simulated data refer to $\Vb^d(t)$ and $\fb(\widehat{\Xb}(t),\thb)$ respectively.  That is what $Y^d$ and $Y^s$ will refer to for the rest of the paper.

\subsection{Algorithmic settings} \label{section:Algorithmic settings}

The success of ABC-SMC algorithm both in terms of computational complexity and quality of the solution depends on the choice of the $\epsilon_\tau$ schedule and the perturbation kernel $K_\tau$. In this section we will briefly describe how we have chosen these two algorithmic settings. A detailed discussion concerning the effects of these settings can be found in \cite{Filippi2013}.  
\subsubsection{Tolerance schedule} \label{section:Tolerance schedule}
Until recently, tolerance values were manually tuned in practice based on prior empirical knowledge about the model. An adaptive choice of the tolerance values has been proposed in \cite{Moral2011} and \cite{Drovandi2011}. In an adaptive schedule the value of the tolerance $\epsilon_\tau$ is chosen as the $\alpha$-th quantile, where $0\leq\alpha \leq 1$ of the distances between the observed data $Y^d$ and simulated data $Y_{\tau-1}^s$ generated at the previous algorithmic time.\\

\subsubsection{Perturbation kernel} \label{section:Perturbation kernel}

Perturbation kernels hold the key to the acceptance rates in ABC-SMC and the speed of the algorithm as exploited in \cite{Filippi2013}.  Perturbation kernels can be broadly divided into two categories: a component-wise perturbation kernel and a multivariate perturbation kernel.  In a component-wise perturbation kernel $\thb\sim\mcal{N}(\thb,\Sigma_{\tau})$ where $\Sigma_{\tau}$ is a diagonal covariance matrix whose diagonal entries $\sigma_{\tau, j}^{2}$ $j=1,\ldots,d$ are chosen adaptively according to the previous population labelled by $\tau-1$ \cite{Beaumont2009,Didelot2011,Filippi2013}. 

A component-wise perturbation kernel is, by construction, unable to generate particles with correlated components; therefore, for models with strongly correlated parameters the ABC-SMC sample generator will not be able to reflect the structure of the posterior and the acceptance rate will be low. Thus, in order to capture such correlations the particles can be perturbed according to a multivariate normal distribution with a non-diagonal covariance matrix $\Sigma_{\tau}$ that depends on the covariance of the particles as reflected in the population in the previous sequential step $(\tau-1)$ \cite{Filippi2013}. Furthermore, a multivariate perturbation kernel operating on a subset of size $N'$ of the $N$ particles (a local kernel) was also shown \cite{Filippi2013} to produce a noticeable improvement in the acceptance rate. In order to define this kernel we will introduce some notation. Let $Y_{\tau}^{s(i)}$ denote the simulated data generated from $\mth$ with particle $\theta\leftarrow\theta_{\tau}^{(i)}$ , $i=1,\ldots,N$ from a population of size $N$ at algorithmic time $\tau$. The corresponding importance weights are denoted as $w_{\tau}^{(i)}$. We collect all such particles (along with the weights) from algorithmic time $\tau-1$ for which $Y_{\tau}^{s(i)}$ is not only within distance $\epsilon_{\tau-1}$ of the observed data $Y^d$ but also within distance $\epsilon_{\tau}$ of it.  We denote such particles as $\tilde{\theta}_{\tau-1}^{(j)}$:
\begin{equation}
\left\{\tilde{\theta}_{\tau-1}^{(j)}\right\}_{1\leq j\leq N'}=\left\{\theta_{\tau-1}^{(i)}\vert \Delta(Y^d,Y_{\tau-1}^{s(i)})\leq \epsilon_\tau,1\leq i\leq N\right\},
\end{equation}
with associated normalised weights $\tilde{w}_{\tau-1}^{(j)}\triangleq({w_{\tau-1}^{(j)}}/{\bar{w}})$, with $\bar{w}\triangleq\sum_j{w}_{\tau-1}^{(j)}$.

Having defined the pairs $\left(\tilde{\theta}_{\tau-1}^{(j)}, \tilde{w}_{\tau-1}^{(j)}\right)$ we can now use a multivariate normal distribution $\mcal{N}(\theta_{\tau-1}^{(i)},\Sigma_{\tau}^{i})$ , with a local covariance $\Sigma_{\tau}^{i}$ (termed the optimal local covariance in \cite{Filippi2013}), to perturb a particle $\theta_{\tau-1}^{(i)}$, where local refers to particle $i$. This covariance is given by
\begin{equation}
\Sigma_{\tau}^{i}=\sum_{j=1}^{N'}\tilde{w}_{\tau-1}^{(j)}\left(\tilde{\theta}_{\tau-1}^{(j)}-\theta_{\tau-1}^{(i)}\right)\left(\tilde{\theta}_{\tau-1}^{(j)}-\theta_{\tau-1}^{(i)}\right)^T.
\end{equation}
%\begin{equation}
%\begin{aligned}
%\Sigma_{\tau}^{i}&=\sum_{j=1}^{N'}\tilde{w}_{\tau-1}^{(j)}\left(\tilde{\theta}_{\tau-1}^{(j)}-m\right)\left(\tilde{\theta}_{\tau-1}^{(j)}-m\right)^T\\
%&+\left(m-\theta_{\tau-1}^{(i)}\right)\left(m-\theta_{\tau-1}^{(i)}\right)^T,
%\end{aligned}
%\end{equation}
%where the term $m=\sum_{j=1}^{N'}\tilde{w}_{\tau-1}^{(j)}\tilde{\theta}_{\tau-1}^{(j)}$ is introduced through a bias-variance decomposition to eliminate any discrepancy between the mean of the restricted sample $\left\{\tilde{\theta}_{\tau-1}^{(j)}\right\}_{1\leq j\leq N'}$ chosen from the population centred around the particle $\theta_{\tau-1}^{(i)}$.  

In the next section we will implement the GP-ABC-SMC algorithm to infer parameters of some standard non-linear differential equations through some toy examples and in that process we will compare and contrast our GP gradient based approach to that of ABC-SMC algorithm with explicit integration. 

\section{Evaluation of the algorithm on benchmark models} \label{sec:eval} 

To evaluate the GP-ABC-SMC algorithm we have chosen three benchmarking differential equations: The Lotka Volterra predator-prey model \cite{murray2002mathematical}, the Hes1 loop model \cite{Monk2003} and signal transduction cascade model \cite{Vyshemirsky2008},\cite{Barber2014}. Each is a set of non-linear differential equations modelling biological systems and show non-trivial dynamical phenomenon such as limit cycle oscillations and non-stationary time evolution. For all these examples we have used the distance function given by (\ref{eq:distance}) and have run the ABC-SMC algorithm with explicit integration using a component-wise univariate normal kernel (ABC-SMC-Comp) \cite{Toni2009} as well as a multivariate normal kernel with the optimal local covariance matrix (ABC-SMC-OLCM) \cite{Filippi2013}. For our proposed GP-ABC-SMC we have also used both the aforementioned perturbation kernels. We refer these as the GP-ABC-SMC and GP-ABC-OLCM respectively. We believe a comparison between these four variants of ABC-SMC is required to capture the difference in speed of execution between the GP based ABC-SMC and the previous approaches reported in \cite{Toni2009,Filippi2013}, while comparing posterior estimates of the parameters. For all the examples presented here, we ran all these variants of ABC-SMC, including the proposed GP based ones, with $N=100$ particles using an adaptive tolerance schedule set to the $\alpha=0.1$ quantile of the distances in the previous populations. The ABC-SMC routines are written in MATLAB and for the GP regressions the GPML package \cite{rasmussen2010gaussian} for MATLAB is used in the predator-prey and Hes1 loop example. For the signal transduction cascade model the GPMat toolbox for MATLAB \texttt{https://github.com/SheffieldML/GPmat} is used which has an implementation of the multi-layer perceptron (MLP) covariance kernel \cite{Barber2014}, required to handle the non-stationarity of some of the state variables. The explicit integrations are carried out using MATLAB's built in ODE and DDE solver routines. 

\subsection{ODE: The predator prey model} \label{subsection:The predator prey model} 
The Lotka Voltera \cite{murray2002mathematical} model depicts an ecological system that is used to describe the interaction between a predator and prey species. This ODE given by
\begin{equation}\label{eq:LOT}
\begin{aligned}
\dot{x}&=\alpha x-xy\\
\dot{y}&=\beta xy-y ,
\end{aligned}
\end{equation}
shows limit cycle behaviour and has been used for benchmarking in \cite{Toni2009,dondelinger2013ode}. $\thb =\left(\alpha,\beta\right)$ is the set of parameters and $\Xb(t)=\left(x(t),y(t)\right)$ is the state vector comprising the concentrations of the predator and the prey species respectively. To create a realistic dataset we generated $11$ uniformly spaced samples between the time interval $(0\leq t\leq 10)$ from the model with parameters $\thb=\left(1,1\right)$ and added random Gaussian noise with zero mean and standard deviation $\sigma=0.5$ to each point. The initial values of the ODE for generating the synthetic data are chosen as $\Xb(t=0)=(1.0, 0.5)$. In order to inspect the consistency of our proposed algorithm we created two more datasets obtained by adding two other realizations of the random noise to the ODE time courses. Thus we have three sets of artificial data (denoted as Dataset $1$, $2$ and $3$), each of which has been corrupted by Gaussian noise with zero mean and standard deviation $\sigma=0.5$ and sampled separately. Note that the GP-ABC-SMC algorithm does not require the estimation of additional nuisance parameters related to the initial values. The time evolution of the state and its derivative is predicted through the GP regression as described in Section \ref{sec:ABC-SMC}. We have used the squared exponential covariance function given by (\ref{eq:SEiso}) for the GP regression in this example. 

From the synthetic data we perform the task of parameter inference using the four different variants of ABC-SMC discussed in the last section to compare their performance. Both $\alpha$ and $\beta$ are chosen from uniform prior distributions $\mathcal{U}(-10,10)$ in all cases. The number of algorithmic iterations, the value of  $S_{\scriptstyle MC}$ is set to $S_{\scriptstyle MC}=6$ for the ABC-SMC-Comp and ABC-SMC-OLCM while it is set to $S_{\scriptstyle MC}=5$ for GP-ABC-SMC and GP-ABC-OLCM. The values differ because we have chosen these on the criteria of minimum number of adaptive iteration required for estimating a reliable posterior distribution. As the ABC-SMC with integration and the GP based variants operate on different spaces thus setting same values for $S_{\scriptstyle MC}$ does not produce comparable results. The specific values of $S_{\scriptstyle MC}$ for this and subsequent examples are chosen on the basis of multiple trials of all the four ABC-SMC algorithm on each of the datasets.
The resulting parameter estimates are listed in Table 1 and the evaluation of the performance in Table 2 (left). We show in Table 1 the mean with $95\% $ confidence intervals of the last population of parameters, approximating the marginal posterior, for each variants of the ABC-SMC. Note that the run-time of the GP-ABC-SMC and the GP-ABC-OLCM algorithms is the sum of the run-time of the ABC and the GP regression (including the estimation of covariance hyperparameters). The value of $\sigma$ is estimated as part of the GP regression. These estimated values are $\sigma=\left\{0.4752,0.8090\right\} $, $\sigma=\left\{0.6219,0.3940\right\}$ and $\sigma=\left\{0.6432,0.4592\right\}$ for the dataset $1$, $2$ and $3$ respectively.
The results show that the GP-ABC-SMC and the GP-ABC-OLCM are considerably faster than the other two variants while producing similar results in terms of the mean of the estimates. However, the variants of ABC-SMC with explicit integration produces narrower confidence intervals compared to the GP variants. Interestingly the GP variants generate comparatively fewer number of samples, as evident from Table 2 (left), indicating higher parameter sensitivities in the derivative space. Also note that the variants of ABC-SMC with a local multivariate kernel generate fewer particles compared to the ones with univariate perturbation kernels, reducing the run-times. Since dimensionality of the local covariance is very small (equal to the ODE parameter space) much less effort is required in its computation than generating simulated data, even in the derivative space.
\begin{table*}
  \centering
	  \caption{Estimated parameters of the Lotka Volterra predator-prey model denoted by the mean of the particles of the final population and $95\%$ confidence intervals for datasets $1$-$3$ respectively in each row. }
  \label{Table:paramLOT}
  \label{Table:paramLOT}
  \begin{tabular}{l*{5}{c}r}
	\toprule
 Parameters & True value & ABC-SMC-Comp & ABC-SMC-OLCM & GP-ABC-SMC & GP-ABC-OLCM\\
\midrule
\multirow{3}{*}{$\alpha$ }&\multirow{3}{*}{$1$}  & $1.0688 \pm 0.0021$ & $1.0722 \pm 0.0032$ & 	$1.0373 \pm 0.0080$ & $1.0356 \pm 0.0070$\\
			&        & $1.0389 \pm 0.0062$ & 			$1.0305 \pm 0.0054$ & 	$1.0843 \pm 0.0092$ & $1.0718 \pm 0.0102$\\
			&        & $1.0293 \pm 0.0157$ & 			$1.0421 \pm 0.0073$ & 	$1.0103 \pm 0.0172$ & $1.0144 \pm 0.0139$\\
\hline
\multirow{3}{*}{$\beta$ }&\multirow{3}{*}{$1$} & $0.9698 \pm 0.0025$ & 	$0.9763 \pm 0.0033$  &  $0.9567 \pm 0.0083$ & $0.9540 \pm 0.0067$\\
			&        & $0.9749 \pm 0.0093$ & 			$0.9966 \pm 0.0082$ & 	$0.9846 \pm 0.0080$ & $0.9760 \pm 0.0083$\\
			&        & $1.0103 \pm 0.0206$ & 			$0.9924 \pm 0.0095$ & 	$0.9887 \pm 0.0154$ & $0.9906 \pm 0.0126$\\
\bottomrule
\end{tabular}
\end{table*}
\begin{table*}
  \centering
	  \caption{Run-time and the ratio of total number of particles accepted to that of generated for the four ABC-SMC algorithms when applied to the three artificial datasets pertaining to the Lotka Volterra predator-prey model (left) and Hes1 model (right). The values for run-time are rounded to nearest integers. }
  \label{Table:paramLOT}
  \begin{tabular}{| c | c | c |}
\hline
 Algorithms & Run-time (seconds) & Accept/Generate\\
\hline
\multirow{3}{*}{ABC-SMC-Comp}& 	$397$ & $700/14737$\\
			&        	$477$ & $600/12294$\\
			&        $516$ & $500/13381$\\
\hline
\multirow{3}{*}{ABC-SMC-OLCM}& 	$184$ & $800/7846$\\
			&        	$221$ & $700/6369$\\
			&        $212$ & $600/6086$\\
\hline
\multirow{3}{*}{GP-ABC-SMC}& 	$25$ & $500/7650$\\
			&        	$26 $ & $500/7547$\\
			&        $26 $ & $500/7642$\\
\hline
\multirow{3}{*}{GP-ABC-OLCM}& 	$21$ & $500/4655$\\
			&        	$20$ & $500/4316$\\
			&        $16$ & $500/3193$\\
\hline
\end{tabular}
  \begin{tabular}{| c | c | c |}
\hline
 Algorithms & Run-time (seconds) & Accept/Generate\\
\hline
\multirow{3}{*}{ABC-SMC-Comp}& 	$106980$ & $1300/763045$\\
			&        	$840330$ & $1600/8026943$\\
			&        $201710$ & $1600/1656197$\\
\hline
\multirow{3}{*}{ABC-SMC-OLCM}& 	$5496$ & $1300/31342$\\
			&        	$8399$ & $1500/50968$\\
			&        $5999$ & $1400/36519$\\
\hline
\multirow{3}{*}{GP-ABC-SMC}& 	$30$ & $1100/31439$\\
			&        	$38 $ & $1000/38911$\\
			&        $32 $ & $1000/36521$\\
\hline
\multirow{3}{*}{GP-ABC-OLCM}& 	$18$ & $1000/7387$\\
			&        	$18$ & $1000/8183$\\
			&        $17$ & $1000/7969$\\
\hline
\end{tabular}\end{table*} 
\subsection{DDE: The Hes1 model} \label{subsection:Hes $1$ model} 
Our proposed algorithm is also able to estimate parameters of delay differential equations. The Hes1 model system is used in systems biology to provide a simplified account of the oscillatory behaviour of the concentrations $(\mu(t),p(t))$ of a species of mRNA and its corresponding protein. The model,  introduced in \cite{Monk2003}, is described by the following delay differential equations:
\begin{equation}\label{eq:Hes1}
\begin{aligned}
\dot{\mu}&=\frac{1}{1+(p(t-t_d)/p_0)^{n}}-\mu_m\mu\\
\dot{p}&=\mu-\mu_p p,
\end{aligned}
\end{equation}
where the parameters $\mu_m$ and $\mu_p$ are decay rates, $p_0$ is the repression threshold, $n$ is the Hill coefficient and $t_d$ is the time delay. We generated data from the above model with parameters $\mu_m=0.03$, $\mu_p=0.03$, $p_0=100$ and $t_d=25$ and initial conditions $\mu(t_0)=3$ and $p(t_0)=3$ for the concentrations between the interval $(0\leq t\leq 300)$ with uniform spacing of $\Delta t=2$ by numerically solving the DDE. $n$ is fixed at a value of $5$ \cite{Monk2003}. We estimated the standard deviations $\sigma_\mu= 6.0020$ and $\sigma_p=121.7670$ of the generated data, for each of the concentrations $\mu(t)$ and $p(t)$. We then added noise, with standard deviation set to $0.1$ times these estimated standard deviations $\sigma_\mu$ and $\sigma_p$, to the data to create the artificial datasets. As in the previous example we created three datasets in a similar fashion.

For comparison of performance of the four methods in the parameter estimation task, we keep the same algorithmic settings, as well as the same covariance function for the GP regression as in the previously example.  Unlike the ODE case where our algorithm does not need to guess the initial state values, it does need a history function for $\Xb(t\leq 0)$ for DDEs in order to work.  In most practical cases the initial history function is taken as a constant function.  Thus in order to make our algorithm work, we shifted the first element of the estimated state evolution backward in time to create the history function.  The four variants of ABC-SMC having the same settings as before, are applied to this artificial dataset.  We chose uniform priors for each of the parameters: $\mu_m\sim\mathcal{U}(-2,2)$, $\mu_p\sim\mathcal{U}(-2,2)$, $p_o\sim\mathcal{U}(0,200)$ and $t_d\sim\mathcal{U}(0,50)$. The number of iterations are chosen as $S_{\scriptstyle MC}=14$ while running ABC-SMC-Comp and ABC-SMC-OLCM. For GP-ABC-SMC and GP-ABC-OLCM this is chosen as $S_{\scriptstyle MC}=9$. As in the previous example these $S_{\scriptstyle MC}$ values are found through multiple trials of each of the algorithms on these datasets and inspecting the quality of the posterior estimates. 
The results are listed in Table 3 and Table 2 (right). In this example we see a huge speedup while using our proposed GP-ABC-SMC and GP-ABC-OLCM algorithms, demonstrating the benefits of this approach. As in the previous example we noticed higher acceptance rates (fewer generated particles) for the GP variants of ABC-SMC. The noise is estimated as $\sigma_\mu=6.8080, \sigma_p=128.6910$, $\sigma_\mu=6.9280, \sigma_p=123.2920$ and $\sigma_\mu=6.1220, \sigma_p=128.1290$ for the dataset $1$, $2$ and $3$ respectively. Furthermore, from Table 3 it is apparent that although the means of the parameters have similar estimates, their corresponding confidence intervals are different between the proposed GP variants and the original ABC-SMC algorithms.  However it should be considered that for GP-ABC-SMC (with both the perturbation kernels) no knowledge of the initial history function is required. Thus in a practical setting we believe a GP based ABC-SMC algorithm is the optimal choice among these four methods for parameter estimation in DDEs.  
\begin{table*}
  \centering
	  \caption{Estimated parameters of the Hes1 loop model. }
  \label{Table:paramLOT}
  \begin{tabular}{l*{5}{c}r}
	\toprule
 Parameters & True value & ABC-SMC-Comp & ABC-SMC-OLCM & GP-ABC-SMC & GP-ABC-OLCM\\
\midrule
\multirow{3}{*}{$\mu_m$ }&\multirow{3}{*}{$0.03$}  & $0.0307 \pm 2.1608 \times 10^{-4}$ & $0.0305 \pm 2.9381 \times 10^{-4}$ & 	$0.0295 \pm 1.3929 \times 10^{-4}$ & $0.0293 \pm 1.1127 \times 10^{-4}$\\
			&        & $0.0341 \pm 6.5563 \times 10^{-5}$ & $0.0342 \pm 4.3428 \times 10^{-5}$ & 	$0.0304 \pm 2.6599 \times 10^{-4}$ & $0.0302 \pm 2.3190 \times 10^{-4}$\\
			&        & $0.0336 \pm 4.1649 \times 10^{-5}$ & $0.0336 \pm 8.5782 \times 10^{-5}$ & 	$0.0291 \pm 1.8721 \times 10^{-4}$ & $0.0292 \pm 1.8657 \times 10^{-4}$\\
\hline
\multirow{3}{*}{$\mu_p$ }&\multirow{3}{*}{$0.03$}  & $0.0294 \pm 2.0457 \times 10^{-4}$ & $0.0297 \pm 2.8397 \times 10^{-4}$ & 	$0.0300 \pm 3.8592 \times 10^{-6}$ & $0.0300 \pm 2.8604 \times 10^{-6}$\\
			&        & $0.0267 \pm 4.4742 \times 10^{-5}$ & $0.0267 \pm 2.9662 \times 10^{-5}$ & 	$0.0300 \pm 4.1123 \times 10^{-6}$ & $0.0300 \pm 3.0506 \times 10^{-6}$\\
			&        & $0.0268 \pm 2.9233 \times 10^{-5}$ & $0.0268 \pm 5.9925 \times 10^{-5}$ & 	$0.0297 \pm 3.5370 \times 10^{-6}$ & $0.0297 \pm 3.6808 \times 10^{-6}$\\
\hline
\multirow{3}{*}{$p_0$ }&\multirow{3}{*}{$100$}  & $99.4130 \pm 0.0537$ & $99.4518 \pm 0.0697$ & 	$99.5991 \pm 0.2946 $ & $99.6997 \pm 0.2058$\\
			&        & $102.1872 \pm 0.0362$ & $102.2306 \pm 0.0281$ & 	$100.8624 \pm 0.2628 $ & $100.8624 \pm 0.2302$\\
			&        & $101.2097 \pm 0.0279$ & $101.2549 \pm 0.0495$ & 	$100.0403 \pm 0.2796 $ & $100.0593 \pm 0.2459$\\
\hline
\multirow{3}{*}{$t_d$ }&\multirow{3}{*}{$100$}  & $25.1318 \pm 0.0109$ & $25.1580 \pm 0.0151$ & 	$25.0496 \pm 0.1139 $ & $25.0502 \pm 0.0871$\\
			&        & $25.2317 \pm 0.081$ & $25.2428 \pm 0.0056$ & 	$25.9357 \pm 0.2031 $ & $25.6215 \pm 0.1589$\\
			&        & $25.0730 \pm 0.0055$ & $25.0714 \pm 0.0107$ & 	$25.3187 \pm 0.1414 $ & $25.4469 \pm 0.1519$\\

\bottomrule
\end{tabular}\end{table*} 
\subsection{ABC variability} \label{subsection:The predator prey model} 
\begin{figure}[!htb]
  \centering
  \subfigure[Distribution of the mean for Lotka Volterra across $50$ runs of GP-ABC-SMC and GP-ABC-OLCM.]{
    \includegraphics[width=7cm]{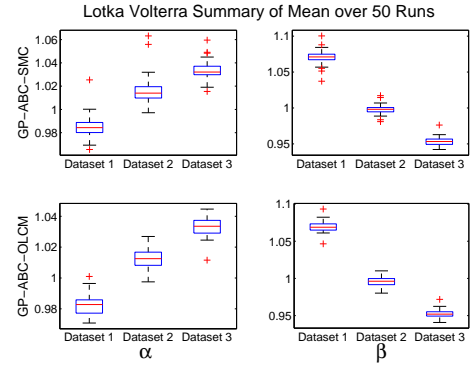}
    \label{Figure:Lotka:left}
  }
	\subfigure[Distribution of the variance for Lotka Volterra across $50$ runs of GP-ABC-SMC and GP-ABC-OLCM.]{
    \includegraphics[width=7cm]{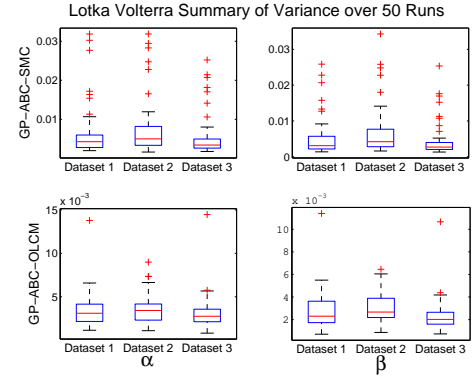}
    \label{Figure:Lotka:right}
  }
  \caption{The boxplots represent the distributions of the mean and variances (across $50$ runs) of the final population representing the marginal approximate posterior parameter distributions learnt by the GP-ABC-SMC and the GP-ABC-OLCM from the three artificial datasets of Lotka Volterra model. }
  \label{Figure:LotkaSummary}
\end{figure}

\begin{figure}[!htb]
  \centering
  \subfigure[Distribution of the mean for Hes1 across $50$ runs of GP-ABC-SMC and GP-ABC-OLCM.]{
    \includegraphics[width=8.1cm]{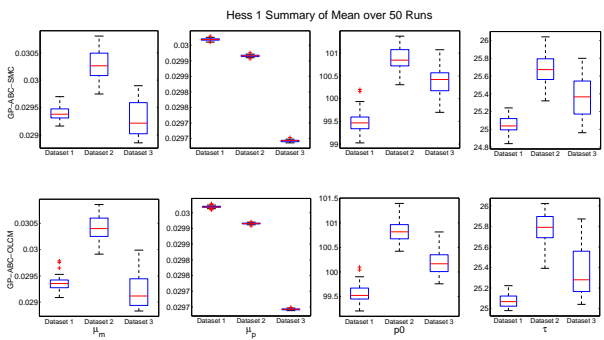}
    \label{Figure:HesSummary:left}
  }
	\subfigure[Distribution of the variance for Hes1 across $50$ runs of GP-ABC-SMC and GP-ABC-OLCM.]{
    \includegraphics[width=8.1cm]{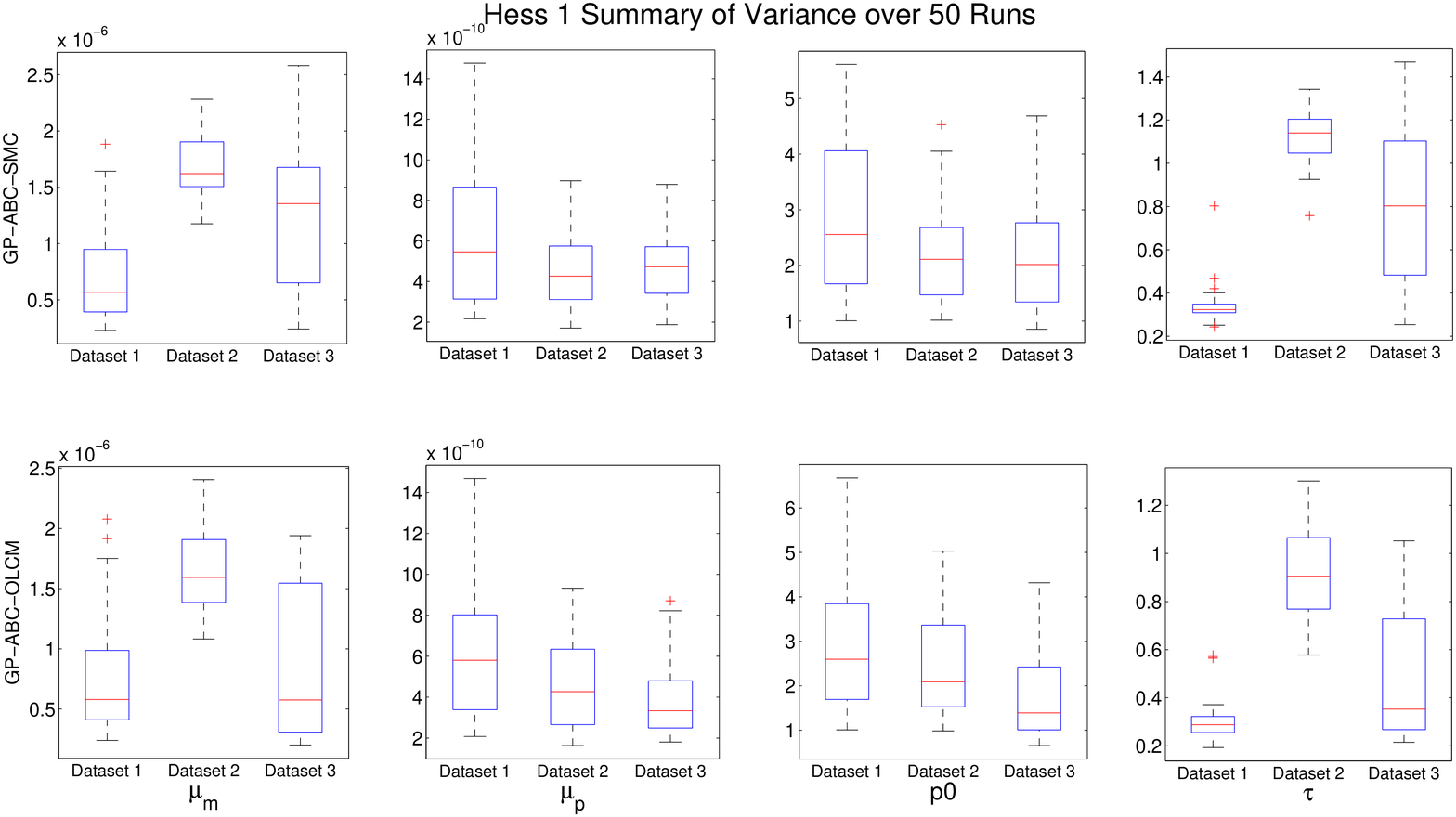}
    \label{Figure:HesSummary:right}
  }
  \caption{Distributions of the mean and variances learnt by the GP-ABC-SMC and the GP-ABC-OLCM from the three artificial datasets of Hes1 model. }
  \label{Figure:HesSummary}  
\end{figure}

Our proposed method comprises of two levels of approximation, one induced through the GP regression and the other one resulting from the approximate inference scheme. Thus in order to check the robustness of our proposed algorithm we repeated the GP-ABC-SMC and GP-ABC-OLCM parameter inference steps for $50$ runs on each of the three artificial datasets for both the Lotka Volterra and Hes1 models. We used the same algorithmic settings and prior distributions as in the previous examples. Fig. \ref{Figure:LotkaSummary}. and Fig. \ref{Figure:HesSummary}. summarizes the distributions of the sample mean and variance (corresponding to the final particle population for each run of of GP-ABC-SMC and GP-ABC-OLCM on the three artificial datasets) across all the $50$ runs on the data from Lotka Volterra and Hes1 respectively.

It is evident from Fig. \ref{Figure:LotkaSummary}. that the GP-ABC-OLCM algorithm produces fewer outliers compared to the GP-ABC-SMC for both the mean and variance estimates. This can be attributed to the local moves in the parameter space caused by the multivariate (OLCM) perturbation kernel. Furthermore, note that the distributions of the variances resulting from each of the algorithms are skewed in a similar manner with the outliers located at the same direction (for both GP variants) from the median. Thus these outliers represent greater variance of posterior distribution of the parameters. However, in more than $90\%$ out of the $50$ runs the moments for both the parameters lie within the inter-quantile range.

\begin{figure}[!htb]
  \centering
  \subfigure[Trajectories of $x(t)$.]{
    \includegraphics[width=8.5cm,height=5cm]{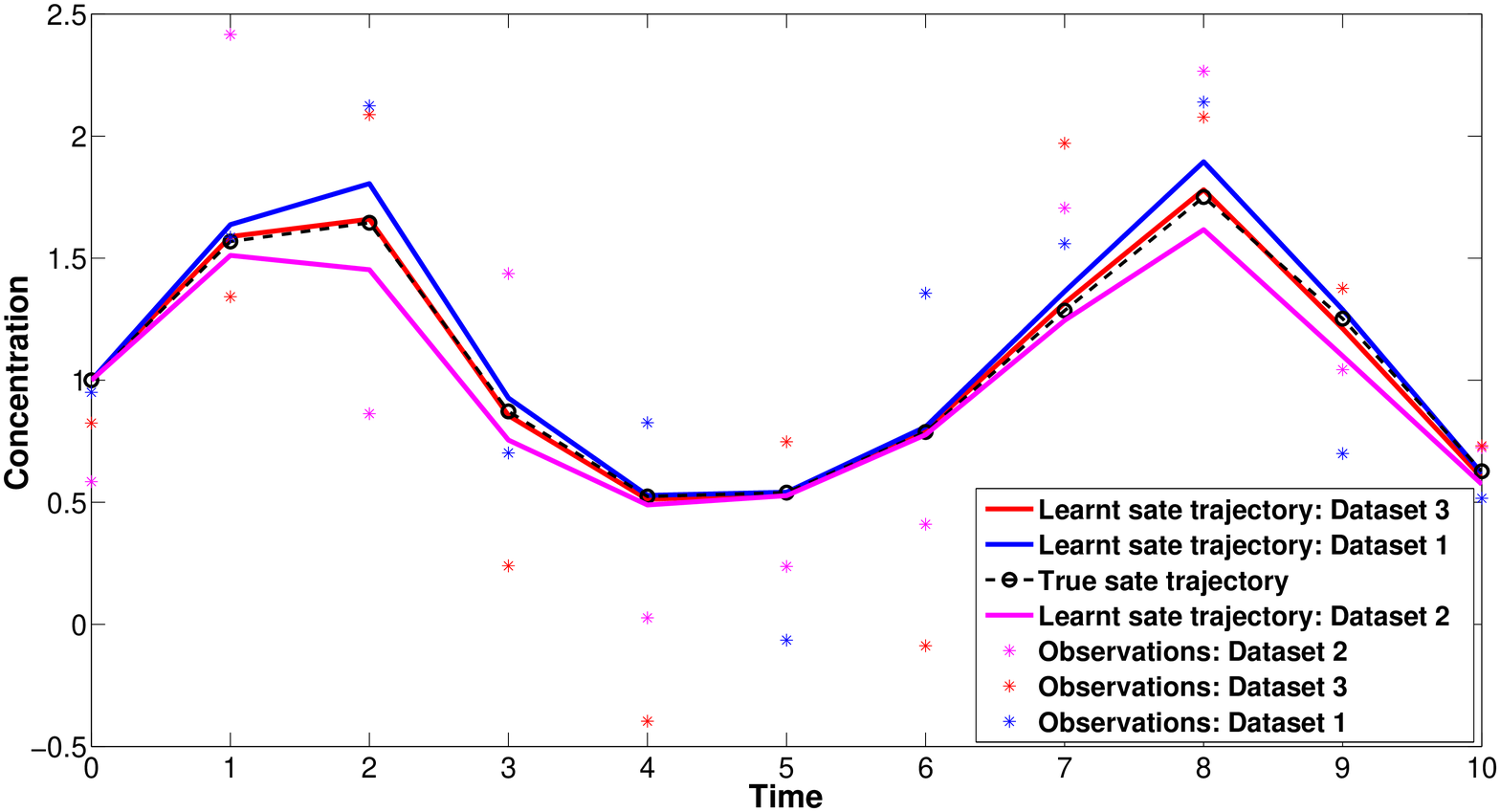}
    \label{Figure:HesSummary:left}
  }
	\subfigure[Trajectories of $y(t)$.]{
    \includegraphics[width=8.5cm,height=5cm]{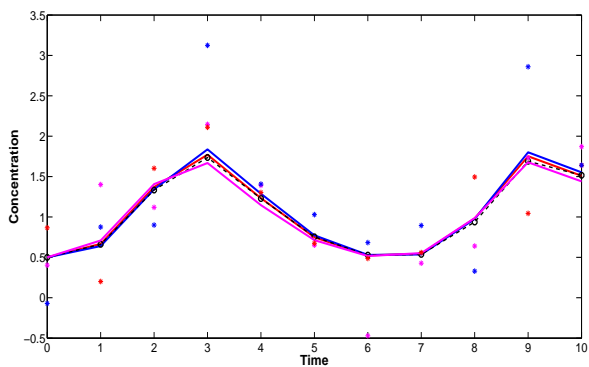}
    \label{Figure:HesSummary:right}
  }
  \caption{Reconstructed and true state trajectories of the Lotka Volterra model. The results corresponding to the three datasets (1, 2, 3) are shown using blue, red and magenta colours respectively. Reconstructed trajectories are represented as curves and observations as stars. The ground truth is the black (dashed and circled)  curve. }
  \label{Figure:LotTraj}  
\end{figure}

\begin{figure}[!htb]
  \centering
  \subfigure[Trajectories of $\mu(t)$.]{
    \includegraphics[width=8.5cm,height=5cm]{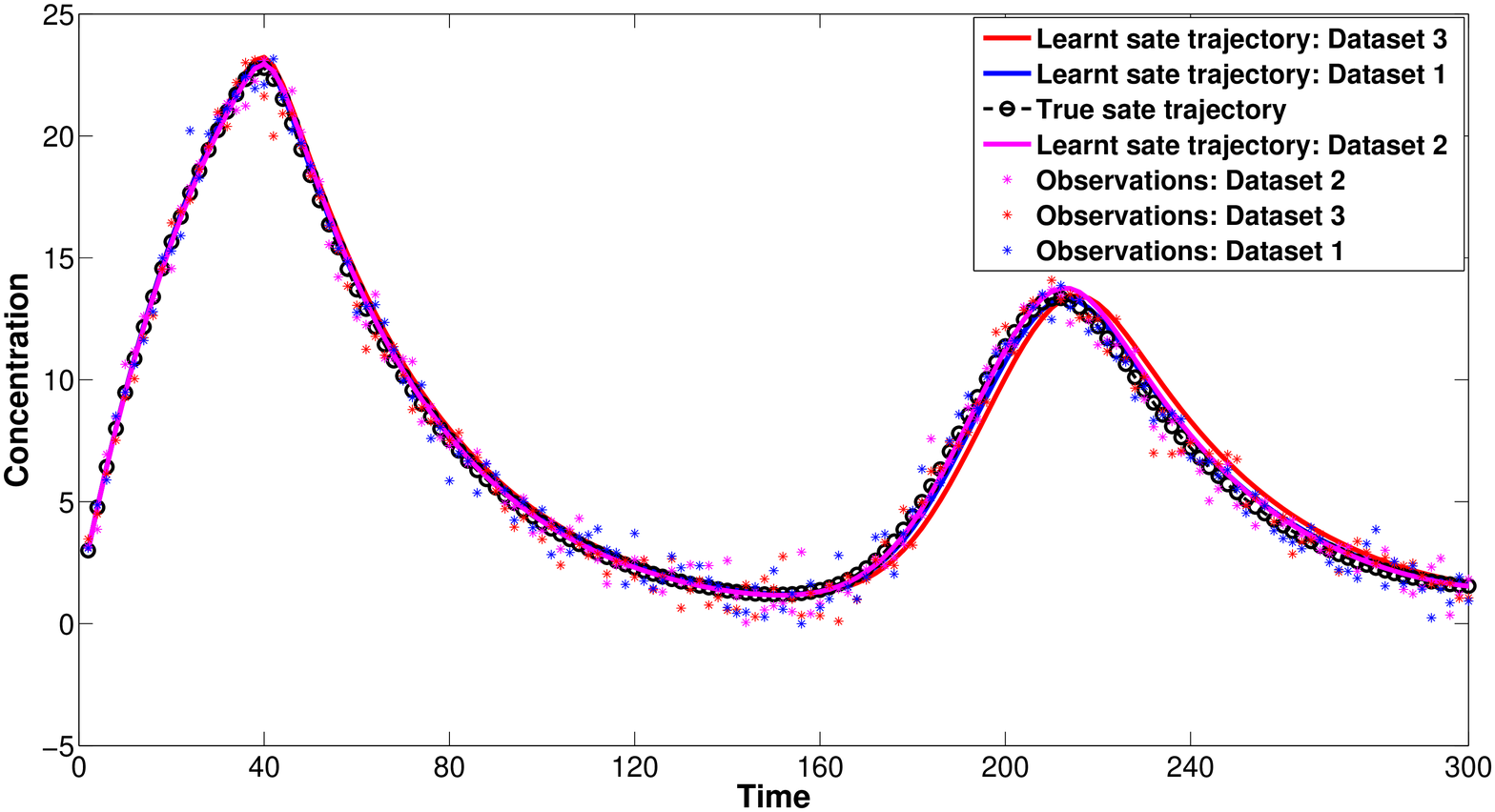}
    \label{Figure:HesSummary:left}
  }
	\subfigure[Trajectories of $p(t)$.]{
    \includegraphics[width=8.5cm,height=5cm]{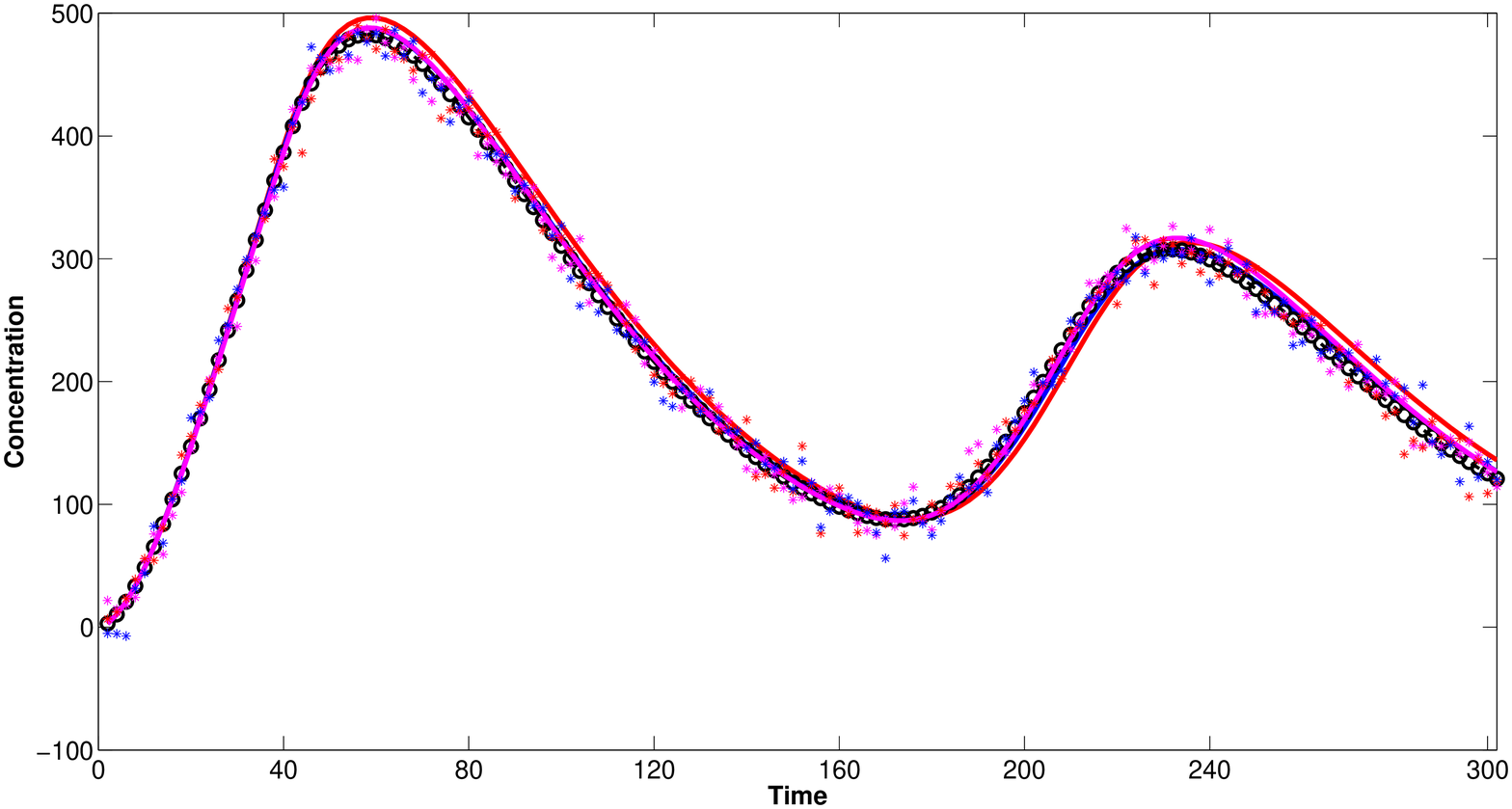}
    \label{Figure:HesSummary:right}
  }
  \caption{Reconstructed and true state trajectories of the Hes1 model. }
  \label{Figure:HesTraj}  
\end{figure}

In case of the Hes1 model it is apparent from Fig. \ref{Figure:HesSummary}. that the distributions are less variable across multiple runs and variants of the algorithms. Moreover, in this case we notice that the distribution of the variances have very few outliers indicating greater accordance among the posteriors learnt after each run of the algorithms. However, it should be noted that an adaptive tolerance schedule results in different (marginally) tolerance values for each new run of the ABC-SMC. Thus some amount of variabilty in the moments  corresponding to different runs is attributed to the differing tolerances. 
Fig. \ref{Figure:LotTraj}. and Fig. \ref{Figure:HesTraj}. show the learnt state trajectories of the Lotka Volterra and Hes1 model compared against the true state trajectories for each of the datasets. The true trajectories correspond to the true parameters and the reconstructed trajectories are generated by solving the Lotka Volterra (\ref{eq:LOT}) and Hes1 (\ref{eq:Hes1}) model equations. While solving (numerically integrating) these differential equations the parameters are taken as the median of the parameters learnt by the GP-ABC-SMC algorithm considering all the $50$ runs. The median value is considered here to reflect the effect of variability (in parameter learning by the GP-ABC-SMC) in reconstructing the dynamics of the considered models.
\subsection{Signal transduction cascade} \label{subsection:Signal transduction cascade} 

We have, so far, used the benchmarking examples to compare our proposed GP-based ABC-SMC approach to others of that ilk that exist in the literature. In this example we will compare the parameter estimation results for the proposed GP based ABC-SMC with other (methods not falling under ABC) recent GP based approximate inference methods for parameter estimation in ODEs. For this purpose we have chosen the signal transduction cascade model \cite{Vyshemirsky2008}. Using this model, a comparison between the competing GP based approaches were reported in \cite{Barber2014}. 
Thus evaluating the proposed GP based ABC-SMC algorithm on this model (with identical settings to those in \cite{Barber2014}) will enable us to draw comparisons with these other methods. This model is described by a 5-dimensional coupled ODEs given by

\begin{equation}\label{eq:LOTBarber}
\begin{aligned}
\frac{d[S]}{dt}&=-k_1[S] -k_2[S][R]+k_3[RS]\\
\frac{d[S_d]}{dt}&=k_1[S]\\
\frac{d[R]}{dt}&=-k_2[S][R]+k_3[RS]+\frac{V[Rpp]}{K_m+[Rpp]}\\
\frac{d[RS]}{dt}&=k_2[S][R]-k_3[RS]-k_4[RS]\\
\frac{d[R_{pp}]}{dt}&=k_4[RS]-\frac{V[Rpp]}{K_m+[Rpp]}\\,
\end{aligned}
\end{equation}
where $\thb =(k_1,k_2,k_3,k_4,V,k_m)$ are the parameters of this model and $\Xb(t)=([S],[S_d],[R],[RS],[R_{pp}])$ are the concentrations of the state variables. Following \cite{Barber2014} we generated data from the model between the time interval $(0\leq t \leq 100)$ with parameters $\thb=(0.07,0.6,0.05,0.3,0.017,0.3)$ and initial values of the state variable $[S]=1$, $[S_d]=0$, $[R]=1$, $[RS]=0$, $[R_{pp}]=0$. We then sampled the data at time $t^L=\{0,1,2,4,5,7,10,15,20,30,40,50,60,80,100\}$ and added random noise with standard deviation $\sigma_{[S]}$, $\sigma_{[S_d]}$, $\sigma_{[R]}$, $\sigma_{[RS]}$, $\sigma_{[R_{pp}]}$ set to $0.1$ for generating the synthetic data. For inferring parameters in this example we apply the GP-ABC-OLCM algorithm from our study with multiple runs, where we found this algorithm to provide a stable and fast inference mechanism. The non-stationarity in the time evolution of the state variables is captured by the MLP covariance function given by
\begin{equation}
\begin{aligned}
k(t,t')& = \sigma^{2}_{kern} \times \\
&\frac{2}{\pi }\text{asin} \left ( \frac{ \sigma_w^2 t^\top t'+\sigma_b^2}{\sqrt{\sigma_w^2t^\top t + \sigma_b^2 + 1}\sqrt{\sigma_w^2 t'^\top t'+ \sigma_b^2 +1}} \right ),
\end{aligned}
\end{equation}
where the kernel variance $\sigma^{2}_{kern}$, the neural network weight variance $\sigma_w^2$, and the bias variance $\sigma_b^2$ are the hyperparameters of the covariance function. The derivative of this kernel with respect to the input time $t$ is given by
\begin{equation}
\frac{\partial k(t,t') }{\partial t}=\frac{\sigma^{2}_{kern}}{\sqrt{1-Z^2}}\frac{\partial Z }{\partial t},
\end{equation}
where 
\begin{equation}
Z=\frac{\sigma_w^2 t^\top t'+\sigma_b^2}{Z_{norm}}
\end{equation}
with $Z_{norm}=\sqrt{\sigma_w^2t^\top t + \sigma_b^2 + 1}\sqrt{\sigma_w^2 t'^\top t'+ \sigma_b^2 +1}$. All the other algorithmic settings were kept the same. The prior distributions are chosen as $k_1\sim\mathcal{U}(0.05,0.09)$, $k_2\sim\mathcal{U}(0.4,0.8)$, $k_3\sim\mathcal{U}(0.03,0.07)$, $k_4\sim\mathcal{U}(0.1,0.5)$, $V\sim\mathcal{U}(0.015,0.0195)$ and $k_m\sim\mathcal{U}(0.1,0.5)$. In this example $S_{\scriptstyle MC}$ is set to $3$.

\begin{figure*}[!htb]
 % \begin{tabular}{cc}
  \centering
  \subfigure[]{
    \includegraphics[width=5.1cm,height=3cm]{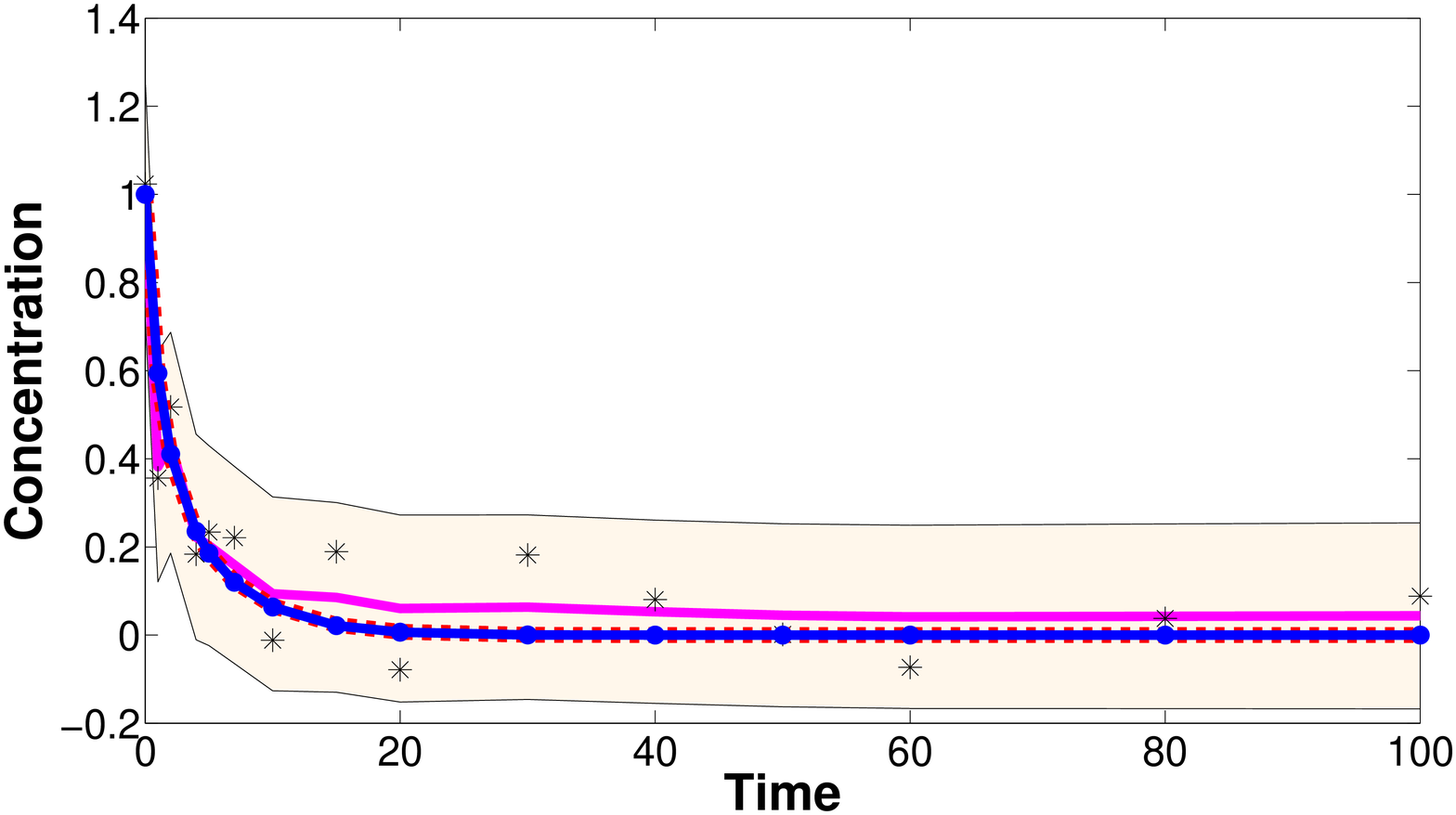}
    \label{Figure:SCSummary:left}
  }  
	\subfigure[]{
    \includegraphics[width=5.1cm,height=3cm]{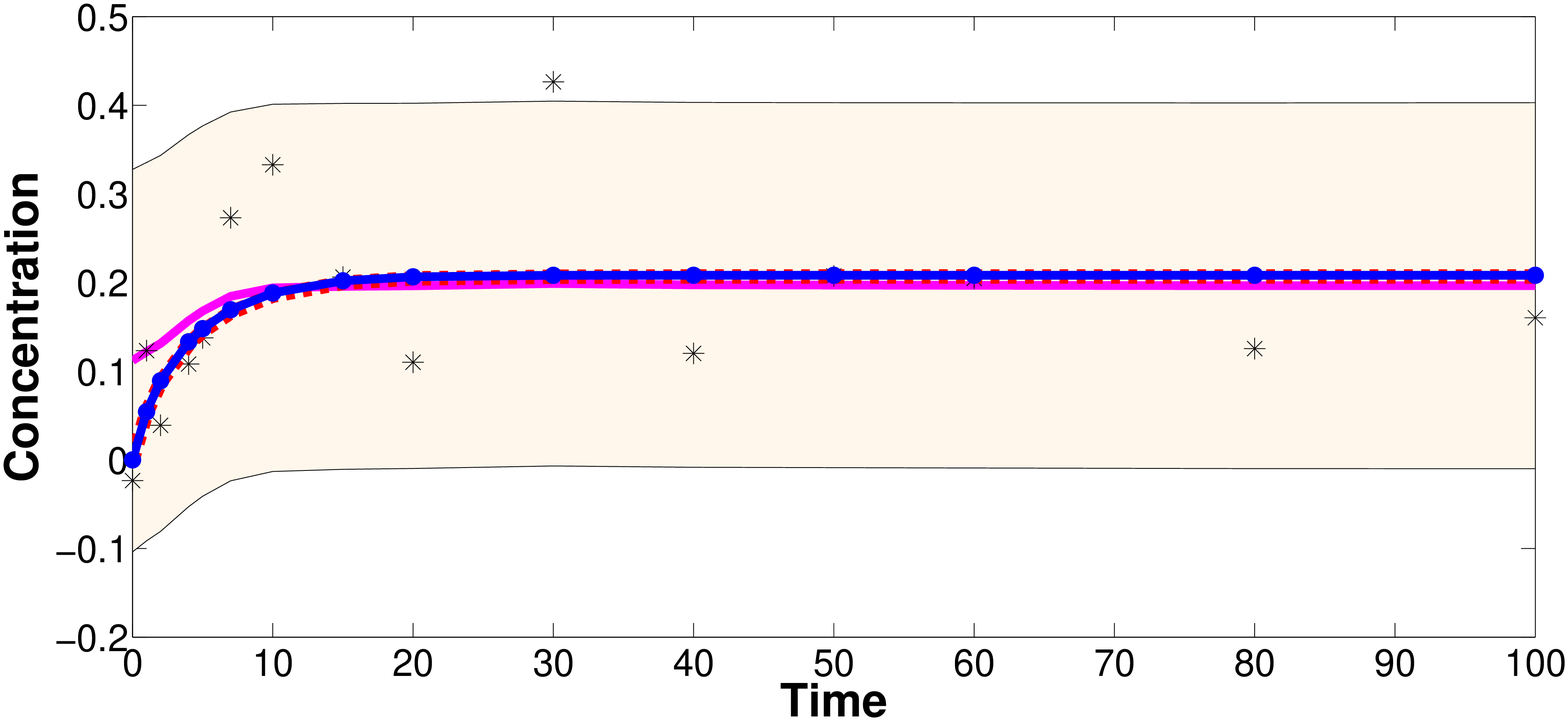}
    \label{Figure:SCSummary:right}
		} 
	\subfigure[]{
    \includegraphics[width=5.1cm,height=3cm]{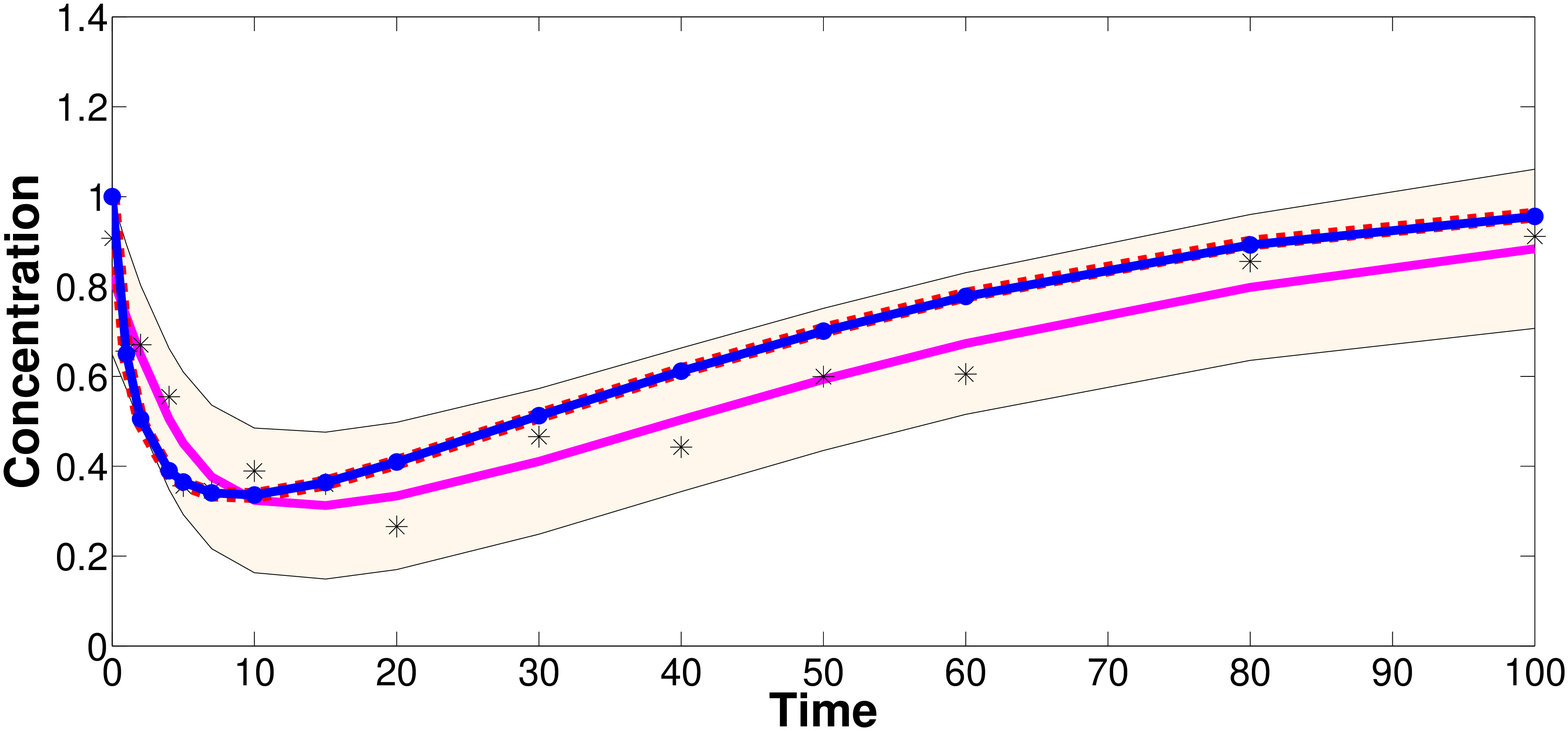}
    \label{Figure:SCSummary:left}
  } 
	\subfigure[]{
    \includegraphics[width=5.1cm,height=3cm]{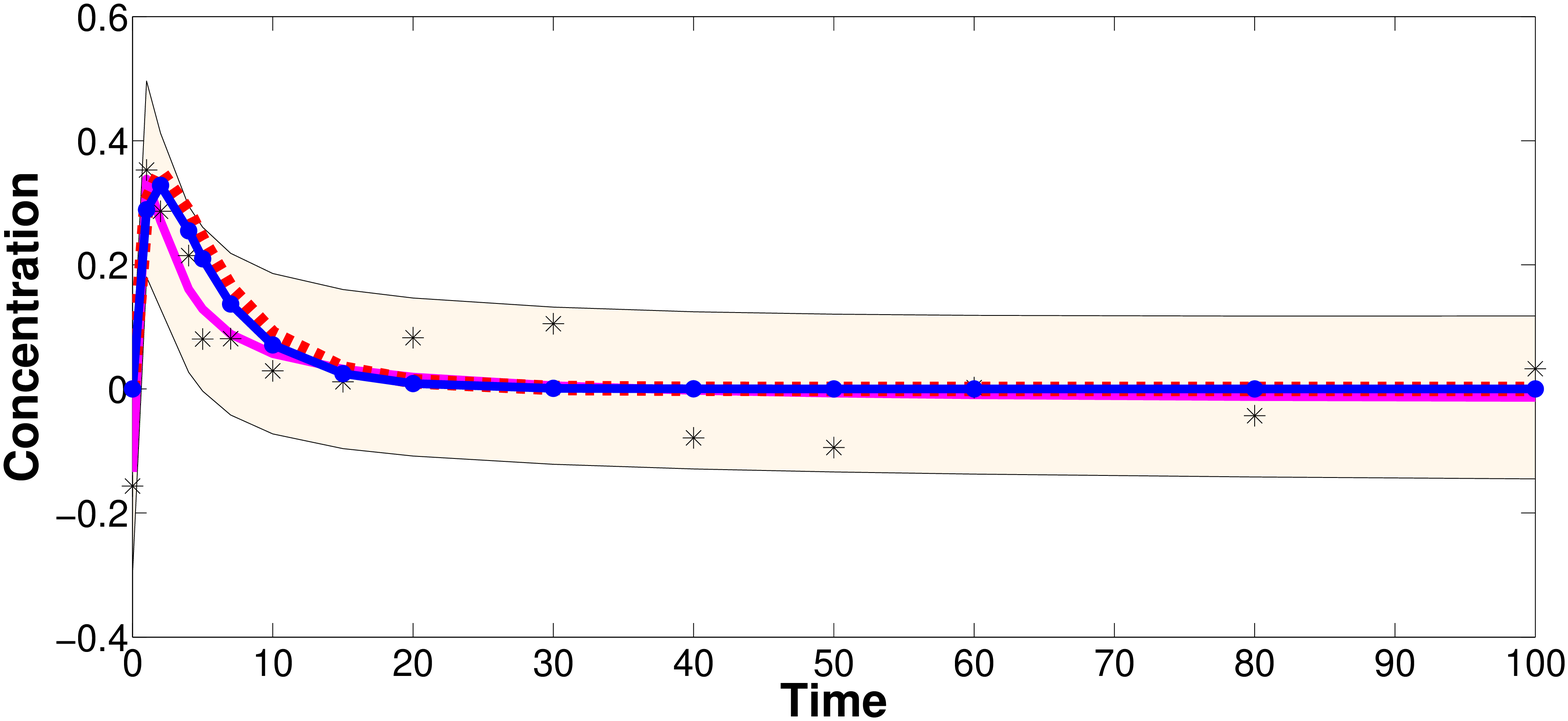}
    \label{Figure:SCSummary:right} 
		}
	\subfigure[]{
    \includegraphics[width=5.1cm,height=3cm]{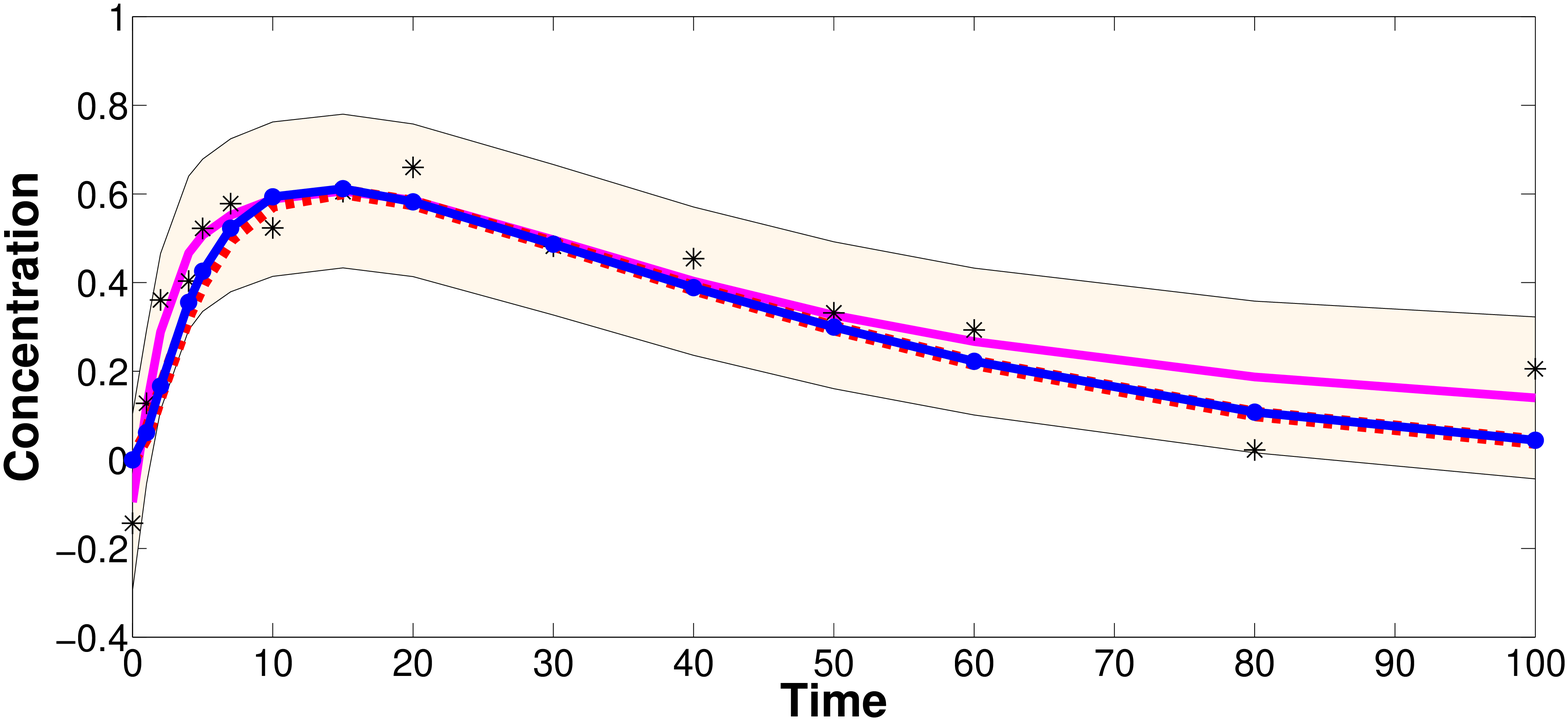}
    \label{Figure:SCSummary:left}
  }
%	\end{tabular}
  \caption{Results of GP-ABC-OLCM for the signal transduction cascade ($([S],[S_d],[R],[RS],[R_{pp}]$ in (a),(b),(c),(d) and (e) respectively). In all the plots observations are the black stars, the true state trajectory is the red (dashed) curve and the blue curve shows the reconstructed trajectory. We have also plotted the GP mean function as the magenta curve and GP variance as the shaded area. The reconstructed trajectory is generated by numerically integrating (\ref{eq:LOTBarber}) with the parameters set to the mean of the posterior distribution estimated by the GP-ABC-OLCM algorithm.}
  \label{Figure:SCSummary}  
\end{figure*}

\begin{figure}[!htb]
  \centering
    \includegraphics[width=8.5cm]{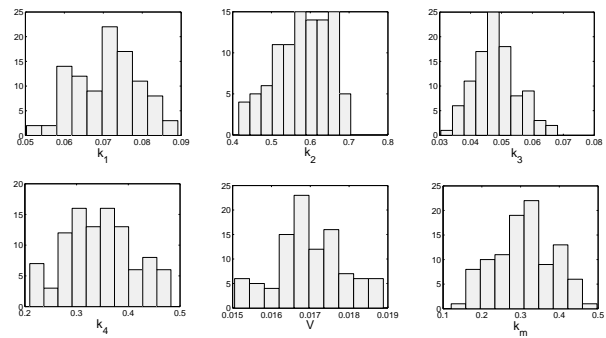}
  \caption{Posterior distributions of the parameters of the signal transduction cascade model learnt by the proposed GP-ABC-OLCM algorithm. }
  \label{Figure:Posterior}  
\end{figure}

The resulting parameter estimates are furnished in Table 4 along with the parameter estimates obtained from other GP based algorithms run on the same model. These algorithms are the \textbf{GP-ODE} method proposed in \cite{Barber2014}, the adaptive gradient matching (\textbf{AGM}) proposed in \cite{dondelinger2013ode} and the gradient matching (\textbf{GM}) proposed in \cite{Calderhead}. The posterior distributions of the parameters as learnt by the GP-ABC-OLCM is shown in Fig. \ref{Figure:Posterior}.
 We have compared the true state trajectories with the reconstructed trajectories in Fig. \ref{Figure:SCSummary}. We generated the reconstructed trajectories by solving (\ref{eq:LOTBarber}) using the mean of the final population of GP-ABC-OLCM, representing the marginal posterior densities of the parameters. The estimated values of the standard deviations are $\sigma_{[S]}=0.0964$, $\sigma_{[S_d]}=0.0818$, $\sigma_{[R]}=0.0707$, $\sigma_{[RS]}=0.0591$ and $\sigma_{[R_{pp}]}=0.0754$.
We avoided the comparison of run-time or acceptance rates as the GP-ABC-OLCM and other GP based algorithms depend on completely different approximate inference scheme. However, GP-ABC-OLCM is significantly faster than the other approaches. The GP-ABC-OLCM finishes the estimation in around $20$ seconds while the other methods were run for $30$ minutes to obtain a properly mixed Markov chain. The ratio of the last two parameters $V/k_m$ \cite{dondelinger2013ode} is a crucial quantity that determines the reconstruction accuracy. GP-ABC-OLCM is able to infer this quantity with the best (based on the estimated posterior means of $V$ and $k_m$) accuracy among all the GP based algorithms. 
\begin{table*}
  \centering
	  \caption{Estimated parameters of the signal transduction cascade by all the GP based approaches including the GP-ABC-OLCM. The estimates for \textbf{GP-ODE}, \textbf{AGM} and \textbf{GM} are taken from \cite{Barber2014}. }
  \label{Table:paramLOT}
  \begin{tabular}{l*{5}{c}r}
	\toprule
 Parameters & True value & GP-ABC-OLCM & \textbf{GP-ODE} & \textbf{AGM} & \textbf{GM}\\
\midrule
$k_1$ & $0.070$ & $0.0708 \pm 0.0086$ & 			$0.0747 \pm 0.0130$ & 	$0.0771 \pm 0.0130$ & $0.0762 \pm 0.0130$\\
$k_2$ & $0.6$ & $0.5806 \pm 0.0706$ & 			$0.6230 \pm 0.1246$  &  $0.5460 \pm 0.1259$ & $0.5632 \pm 0.1256$\\
$k_3$ & $0.05$ & $0.0480 \pm  0.0074$ & 			$0.0530 \pm 0.0135$ & 	$0.0593 \pm 0.0111$ & $0.0594 \pm 0.0115$\\
$k_4$ & $0.3$ & $0.3439 \pm 0.0659$ & 			$0.2960 \pm 0.0281$  & 	$0.3750 \pm 0.0999$ & $0.3754 \pm 0.1051$\\
$V$ & $0.017$ & $0.0170 \pm 0.0009$ & 			$0.0177 \pm 0.0014$ & 	$0.0172 \pm 0.0015$ & $0.0173 \pm 0.0014$\\
$k_m$ & $0.3$ & $0.3110 \pm  0.0774$ & 			$0.4220 \pm 0.0690$  & 	$0.4090 \pm 0.0911$ & $0.4186 \pm 0.0953$\\

\bottomrule
\end{tabular}
\end{table*}


\section{Conclusion}\label{sec:conc}

In this paper we have proposed a method that significantly speeds up the task of parameter inference in comparison with state of the art methods that use ABC and SMC based approaches when applied to several benchmark dynamical system models that are described by ordinary and delay differential equations . We achieve this speed-up by circumventing the need to numerically integrate the differential equations, a task that is repeatedly required in other ABC methods to generate samples from candidate models for comparison with the observed data. The key idea behind our method lies in building on \cite{Calderhead}, \cite{Ramsay2007} and \cite{Barber2014} to work directly with the vector field of the dynamical system, which we model using Gaussian process regression, and thus create a distance function in derivative space for use in the ABC-SMC algorithm as proposed in \cite{Toni2009}. Thus we proposed a modified ABC-SMC algorithm for parameter estimation in ODEs or DDEs.  

We benchmarked the benefits of this approach by evaluating our proposed approach on toy problems where we observed a significant speed-up of the parameter estimation process. We also compared our proposed approach with other GP based methods proposed in recent literature and found that our proposed GP-ABC-SMC (with the local multivariate perturbation kernel) performs significantly faster to obtain similar estimates. Furthermore, improvements of ABC-SMC through perturbation kernels as proposed in \cite{Filippi2013} and has been integrated with our approach to obtain enhanced performance. Thus, our fast approximate inference process can accommodate the useful features of ABC-SMC (as shown in \cite{Toni2009}) such as model selection and sensitivity analysis. Our proposed approach is only limited by the ability of the Gaussian process regression in smoothing the observed time series data while retaining the essential characteristics that are meant to be captured by the dynamical system model.  Thus in those cases where smoothing the experimental data by GP regression introduces artefacts, the GP-ABC-SMC algorithm would produce poor parameter estimates.  As future work we wish to extend the GP-ABC-SMC algorithm for stochastic differential equation by relating the GP regression technique with drift estimation technique in \cite{Andreas2013}.  We will also include models with hidden variables, as the smoothing procedure on observed data can no longer provide complete information.

% or



\ifCLASSOPTIONcompsoc
  \section*{Acknowledgments}
\else
  \section*{Acknowledgment}
\fi

The work reported in this paper was supported by project PLants Employed As SEnsor 
Devices (PLEASED), EC grant agreement number 296582.

\ifCLASSOPTIONcaptionsoff
  \newpage
\fi



%\bibliographystyle{IEEEtran}
%\bibliography{minths}

\end{document}